\documentclass[10pt,twocolumn,letterpaper]{article}

\usepackage{cvpr}      % To produce the REVIEW version
\usepackage{url}
\usepackage{enumitem}
\usepackage{graphicx}
\usepackage{booktabs}
\usepackage{multirow}
\usepackage{multicol}
\usepackage{makecell}
\usepackage{comment}
\usepackage{colortbl}
\usepackage{subcaption}
\usepackage{float}
\usepackage{listings}
\usepackage[dvipsnames]{xcolor}
\usepackage{stfloats}
\usepackage[utf8]{inputenc} % allow utf-8 input
\usepackage[T1]{fontenc}    % use 8-bit T1 fonts
\usepackage{amsfonts}       % blackboard math symbols
\usepackage{nicefrac}       % compact symbols for 1/2, etc.
\usepackage{microtype}      % microtypography
\usepackage{xcolor}         % colors
\usepackage{titletoc}
\definecolor{cvprblue}{rgb}{0.21,0.49,0.74}
\usepackage[pagebackref,breaklinks,colorlinks,allcolors=cvprblue]{hyperref}

\def\paperID{8746} % *** Enter the Paper ID here
\def\confName{CVPR}
\def\confYear{2025}

\title{SeqAfford: Sequential 3D Affordance Reasoning via \\Multimodal Large Language Model}

\author{Chunlin Yu$^{1,\dagger}$
\quad  Hanqing Wang$^{1,\dagger}$
\quad Ye Shi$^1$
\quad Haoyang Luo$^1$
\quad  Sibei Yang$^1$\quad Jingyi Yu$^1$
\quad Jingya Wang$^{1,}$
\thanks{Corresponding author.} \\
$^1$ShanghaiTech University, Shanghai, China\\
{\tt\small \{yuchl,wanghq2024,shiye,luohy2024, yangsb,yujingyi,wangjingya\}@shanghaitech.edu.cn}\\
}

\begin{document}
% \maketitle

\twocolumn[{
\renewcommand\twocolumn[1][]{#1}
\maketitle
\thispagestyle{empty} % no page number for the first page
\begin{center}
    \captionsetup{type=figure}
    \includegraphics[width=0.97\linewidth]{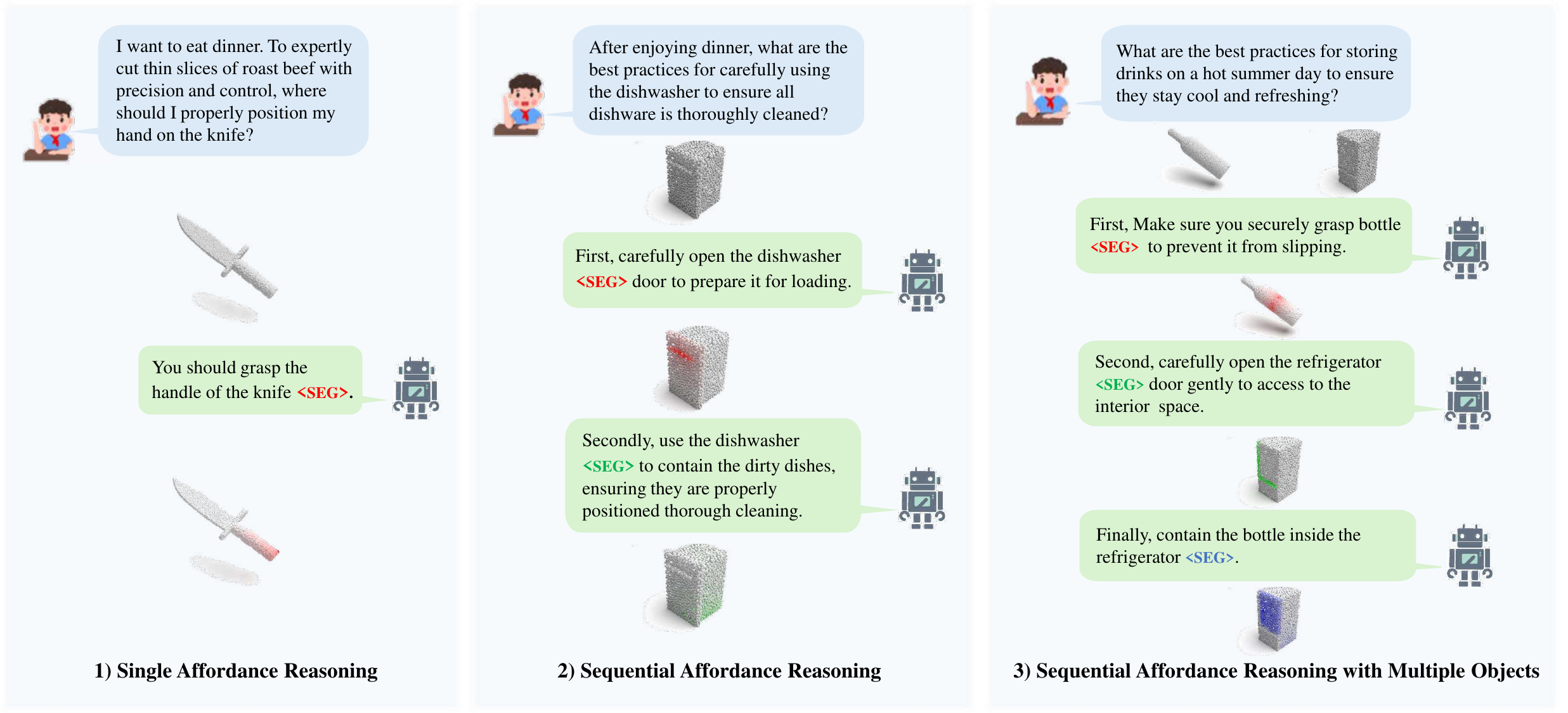}
    \captionof{figure}{\textbf{Sequential 3D affordance reasoning task with different types of interactions.} We introduce SeqAfford, a Multi-Modal Language Model (MLLM) capable of serialized affordance inference implied in human instructions: 1) Single Affordance Reasoning; 2) Sequential Affordance Reasoning; 3) Sequential Affordance Reasoning with Multiple Objects}
\end{center}
\label{figvis}
}]
\def\thefootnote{$\dagger$}\footnotetext{Equal contributions.}
\def\thefootnote{*}\footnotetext{Corresponding author.}

\begin{abstract}
3D affordance segmentation aims to link human instructions to touchable regions of 3D objects for embodied manipulations. Existing efforts typically adhere to single-object, single-affordance paradigms, where each affordance type or explicit instruction strictly corresponds to a specific affordance region and are unable to handle long-horizon tasks. Such a paradigm cannot actively reason about complex user intentions that often imply sequential affordances. In this paper, we introduce the Sequential 3D Affordance Reasoning task, which extends the traditional paradigm by reasoning from cumbersome user intentions and then decomposing them into a series of segmentation maps. Toward this, we construct the first instruction-based affordance segmentation benchmark that includes reasoning over both single and sequential affordances, comprising 180K instruction-point cloud pairs. Based on the benchmark, we propose our model, SeqAfford, to unlock the 3D multi-modal large language model with additional affordance segmentation abilities, which ensures reasoning with world knowledge and fine-grained affordance grounding in a cohesive framework. We further introduce a multi-granular language-point integration module to endow 3D dense prediction. Extensive experimental evaluations show that our model excels over well-established methods and exhibits open-world generalization with sequential reasoning abilities. Project page: \url{https://seq-afford.github.io/}.
\end{abstract}    
\section{Introduction}
Affordance is a crucial lens through which humans and embodied agents interact with various objects of the world . When provided with human instructions, affordance aims to highlight the actionable possibilities of these objects, linking visual perception with manipulation. While 2D affordance~\cite{qian2024affordancellm, li2024one, xuweakly} offers visual cues that suggest potential actions to embodied systems, 3D affordance~\cite{delitzas2024scenefun3d, chu20253d, yang2024lemon, gao2024learning, lu2024geal} provides a more direct and intuitive guidance for executing tasks in the realistic 3D world, and thus solidify the foundation for downstream robot manipulation tasks~\cite{wu2024afforddp, ning2023where2explore, ju2024robo, wu2023learning}.

Previous work on 3D affordances has largely focused on the single-object, single-affordance paradigm, where affordance maps are grounded either in affordance categories~\citep{deng20213d, nguyen2023open} or 2D demonstration images~\citep{yang2023grounding, gao2024learning}. Recently, language models have been employed to pair 3D objects with natural language questions~\citep{li2024laso}, each designed to elicit a specific affordance. For example, the question “How can you go through the door?” can be interpreted by language models like BERT \cite{kenton2019bert} or RoBERTa \cite{liu2019roberta} as referring to the “openable” affordance component (e.g., the door handle). However, regardless of whether the grounding source is an affordance phrase, HOI image, or a natural question, each grounding source generally corresponds to a fixed affordance type.
%While these approaches are successful at linking specific affordances to natural language, they are limited in their ability to understand complex user intentions that require broader world knowledge or reasoning beyond simple affordance mappings 

Current systems can't actively reason complex user intentions, break them down into actionable primitives, or formulate chains of affordances from each primitive. Real-world physical interactions require modeling complex human instructions involving multiple objects and sequential affordances. For example, to 'reheat food in a bowl using the microwave', agents must first grasp the bowl, open the microwave, then contain the bowl inside. Thus, multi-object sequential reasoning is crucial for next-generation affordance systems.

Recently, Large-Language Models (LLMs)  ~\cite{alayrac2022flamingo, zhu2023minigpt, ouyang2022training} have demonstrated exceptional sequential reasoning abilities, ingrained with internalized common-sense knowledge encoded from vast text data corpora. On top of that, the emergence of 3D Multimodal Large-Language Models (MLLMs)  ~\cite{qi2024shapellm, xu2023pointllm} have further expanded their possibilities in understanding various object shapes in the 3D world. However, even the latest 3D MLLMs are not panaceas for reasoning about visual affordances from 3D objects, as they primarily focus on object-centered text generation tasks. This, therefore, highlights a pressing question: \textit{Can we devise a 3D multi-modal large language model to sequentially reason and segment multi-object affordances based on long-horizon human instructions?}

\begin{figure*}[ht]
    \centering

    \includegraphics[width=0.95\textwidth]{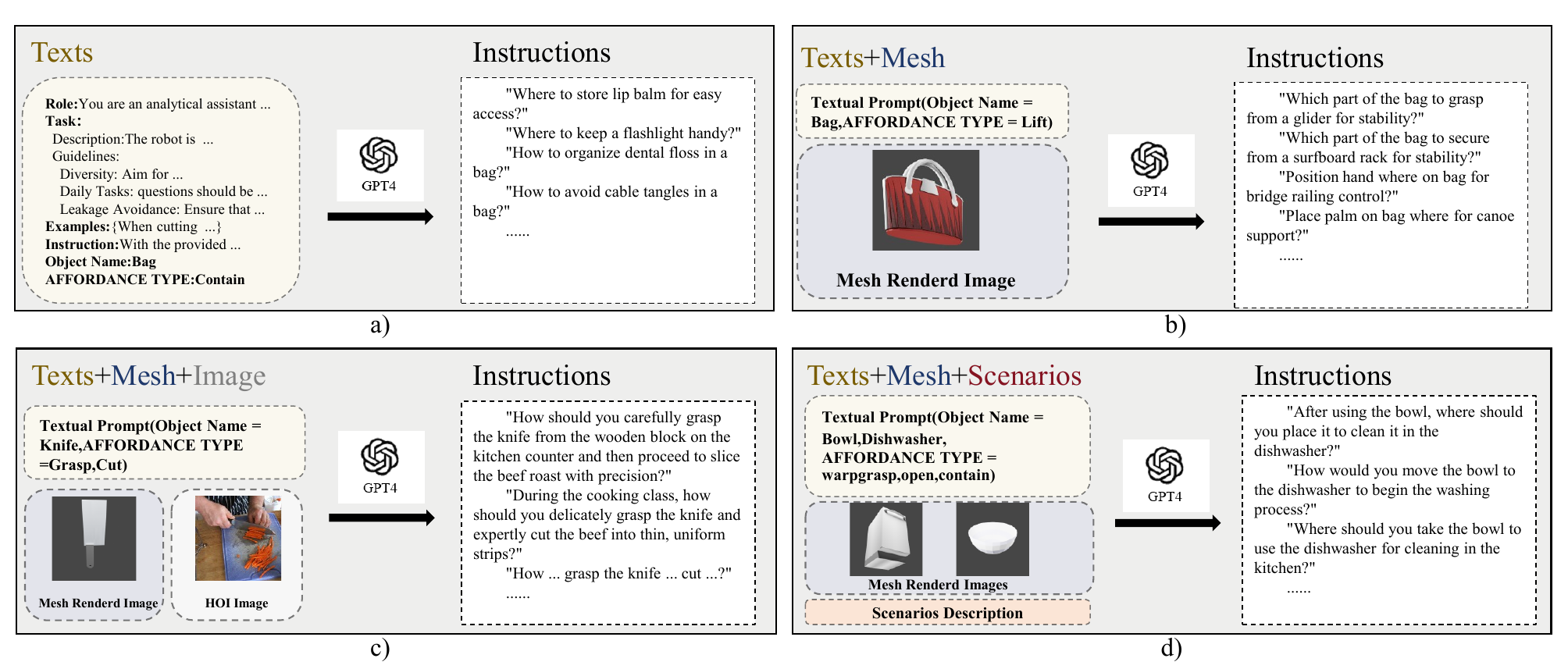}
    \caption{\textbf{Preparing the instructions.} 
     To better utilize the world knowledge of GPT4, We prompt GPT-4o to generate diverse instructions based on 4 types of system prompts containing different modalities as input. Instructions are generated based on input prompts with modalities from a) purely textual affordance type, object name; b) the mesh-rendered image of the object; c) the mesh-rendered image and HOI images that reveal affordances of the object; d) the mesh-rendered image and textual description of the scenario. }
    \label{fig1}
\end{figure*}

In this paper, we introduce a new task called Sequential 3D Affordance Reasoning, designed to narrow the gap with real-world demands. Towards this, we construct a large-scale sequential affordance reasoning benchmark, containing 180K instruction-point cloud pairs. To ensure the diversity of the instruction data, four distinct methods are used for generating instructions. These methods include utilizing immersive 2D HOI images and detailed, colorful mesh renderings to prompt GPT-4o to generate diverse instructions, drawing on its inherent world knowledge, as illustrated in Fig. \ref{fig1}. Supported by this benchmark, we introduce our model SeqAfford, which unlocks the current 3D MLLMs with sequential affordance segmentation abilities. To further bolster the 3D dense prediction and reasoning task, we introduce a multi-granular language-point integration module, where dense point features conditioned on segmentation tokens of the large language model are integrated with sparse point features for subsequent dense prediction tasks. This module not only effectively injects the reasoning results of large language models into the dense point features, but enhances the affordance segmentation task with multi-granular levels of representation. 

To summarize, our contributions are as follows:
\begin{itemize}
\item[$\bullet$] We introduce the Sequential 3D Affordance Reasoning task, which involves the sequential reasoning and segmentation of affordances based on complex human instructions. This paradigm is crucial for the development of next-generation affordance systems. 

\item[$\bullet$] We develop a large-scale sequential affordance reasoning benchmark with 180K instruction-point cloud pairs, serving as a comprehensive resource to advance research in affordance reasoning. 

\item[$\bullet$] We leverage the internalized world knowledge of pre-aligned 3D MLLMs to achieve fine-grained affordance-level multi-target sequential reasoning and explanations in a cohesive framework.
\end{itemize}
\section{Related Work}

\textbf{3D Affordance Segmentation.} With the rapid advancement of embodied AI, research on 3D affordance has increasingly garnered significant attention from both academia and industry. 3D AffordanceNet   \citep{deng20213d}, built on point cloud data from PartNet   \citep{mo2019partnet}, was the first to construct a fine-grained 3D affordance dataset, establishing a benchmark for 3D affordance research. Building on this,  some researchers ~\citep{yang2023grounding,shao2024great} proposed leveraging universal knowledge extracted from 2D human-object interaction (HOI) images to assist in 3D affordance segmentation. However, these methods rely solely on visual information to infer affordances, without considering that embodied agents in the real world communicate with humans through language. Consequently, such methods are limited in their applicability for direct deployment in embodied agents. Recently,  ~\citet{li2024laso} introduced a language-based affordance segmentation task, promoting the integration of natural language and affordance understanding. 

Following this,  ~\citet{chu2024iris} utilized LLMs to locate objects in 2D images and subsequently retrieve corresponding objects from a 3D dataset. Although these approaches leverage the reasoning capabilities of LLMs, they lack the ability to perform joint visual and language alignment, failing to bridge the gap between 2D and 3D modalities. Furthermore, their research assigns each text statement to a single specific affordance, overlooking the complexity of scenarios that often require the coordination of multiple affordances. 
In response to these limitations, we propose the integration of 3D multimodal large language models (MLLMs) into affordance segmentation. This novel approach enables the model to simultaneously comprehend contextual semantics and point cloud data, thereby facilitating affordance reasoning across a wide range of complex scenarios. By incorporating both natural language and 3D data, our model offers a more comprehensive and versatile understanding of affordances, rendering it better suited for real-world applications in embodied AI.

\noindent \textbf{Multimodal Large Language Models.} Large Language Models (LLMs) have achieved remarkable success in processing natural language, and researchers have been working to extend these models' reasoning capabilities into the visual domain. Early efforts \cite{alayrac2022flamingo,liu2024visual,zhu2023minigpt,nasiriany2024rt} have made significant strides by enabling LLMs to process both visual and textual information concurrently, thereby introducing an initial level of human-like multimodal reasoning. However, for many practical applications, such as visual segmentation, these models lack the necessary fine-grained perception required for detailed visual tasks. To address this issue, research efforts \cite{liu2023interngpt,wang2024visionllm,you2023ferret,zhang2023gpt4roi} enable the localization of specific regions within images by encoding spatial coordinates as tokens, improving the models' ability to reason about precise areas within the visual data. 
% such as GPT4RoI~\citep{zhang2023gpt4roi}, VisionLLM~\citep{wang2024visionllm}, InternGPT~\citep{liu2023interngpt}, and Ferret~\citep{you2023ferret} 

% Additionally, models such as LISA   \citep{lai2024lisa} and PixelLM   \citep{ren2024pixellm} improve MLLMs’ capacity to handle pixel-level tasks through the incorporation of specialized tokens and modules like SAM   \citep{kirillov2023segment}, extending the application of MLLMs to more complex visual reasoning tasks.
% 
Building on these 2D MLLM advancements, research has increasingly expanded into the 3D domain. Following the paradigm of LLaVA, PointLLM ~\citep{xu2023pointllm} replaces the vision encoder with a 3D point encoder, enabling the processing of 3D data within the latent space of LLMs and facilitating 3D object understanding. Similarly, ShapeLLM   ~\citep{qi2024shapellm} is built upon the enhanced 3D encoder RECON++, which strengthens geometric understanding through multi-view image distillation. Other models, such as 3D-LLM   ~\citep{hong20233d}, leverage 2D foundational models, like CLIP ViT   ~\citep{radford2021learning}, to process multi-view rendered images of 3D point clouds, thereby integrating the 3D world into LLMs. 
% Furthermore, models like 3D-VLA   \citep{zhen20243d} and ManipLLM   \citep{li2024manipllm} introduce the action modality, enabling embodied agents to navigate and interact within 3D environments in a more dynamic and context-aware manner. Recent works, such as SegPoint   \citep{he2024segpoint}, further improve the adaptability of 3D MLLMs by incorporating specialized tokens for specific downstream tasks.
% 
However, despite these advancements, the majority of existing MLLMs are primarily focused on scene-level and object-level understanding, lacking the ability to recognize and segment fine-grained affordances of 3D objects in diverse semantic contexts. Addressing this limitation, our study aims to endow MLLMs with affordance-aware perception, enabling them to interpret and act upon 3D objects more effectively in context-sensitive scenarios. 
\section{Dataset}
\label{data}
\begin{table*}[ht]
\label{Table2}
\centering

\begin{tabular}{@{}lccccccc@{}}
\toprule
 Method    & \#Sequential  & \#World Knowledge & \#Multi-object & \#Instruction-Point Pairs &  \#Point Cloud      \\
\midrule
3D AffordanceNet  \citep{deng20213d}  & $\times$  & $\times$ & $\times$ & $\times$ & 23k   \\
O2O-Afford  \citep{mo2022o2o} &  $\times$ & $\times$ & $\times$  & $\times$ & 1.7k    \\
Partafford  \citep{xu2022partafford}  & $\times$  & $\times$ & $\times$ & $\times$ & 25k   \\
IAGNet \citep{yang2023grounding} & $\times$ & $\times$  & $\times$  & $\times$ & 7k   \\
LASO  \citep{li2024laso} & $\times$ &$\times$ & $\times$ & 19k & 8.4k       \\
Ours & $\checkmark$  & $\checkmark$ & $\checkmark$   & 180k & 18k  \\
\bottomrule
\end{tabular}

\caption{\textbf{Comparison of Exisiting 3D Affordance Dateset with Ours.} \#Point Cloud and \#Instruction-Point Cloud Pairs denote the number of point clouds and instruction-point cloud pairs, respectively. $\times $  indicates that the dataset does not possess this attribute. 
}
\end{table*}
% You must include your signed IEEE copyright release form when you submit your finished paper.
% We MUST have this form before your paper can be published in the proceedings.

% Please direct any questions to the production editor in charge of these proceedings at the IEEE Computer Society Press:
% \url{https://www.computer.org/about/contact}.
Affordance segmentation involves understanding the operability of objects in various contexts, and the varying complexity of intentions adds significant challenges to the process. Simple instructions typically pertain to the direct usage of an object, such as grasping a cup or opening a door. In contrast, complex instructions may involve multi-step actions or require contextual understanding, such as using a cup for a specific occasion or purpose. To address the challenge of affordance segmentation based on both simple and complex instructions, we constructed a dataset of instruction-point cloud pairs based on 3D AffordanceNet   \citep{deng20213d}, encompassing both simple and complex types of intentions, which includes 162,386 instruction-point cloud pairs in the single affordance segmentation setting and 20,847 pairs in the sequential affordance segmentation setting, comprising a total of 18,371 point cloud instances across 23 object categories. 

\subsection{Dataset Collection}
\textbf{Point Cloud.} Our point cloud data and affordance annotations are entirely sourced from 3D AffordanceNet    \citep{deng20213d}. In the simple instruction setting, we generated five instructions for each affordance of every point cloud instance. For the sequential affordance segmentation setting, we carefully selected point cloud categories that support this configuration and generated corresponding instructions for each affordance sequence combination.

\noindent \textbf{Instruction.} To create instructions, we developed four methods for generating instructions using GPT-4   \citep{achiam2023gpt}. Unlike LASO   \citep{li2024laso} which generates the same texts for all point cloud instances of a specific affordance type for each category, we generate text for each point cloud by utilizing the point cloud instance from 3D AffordanceNet   \citep{deng20213d} to trace back to the Mesh-rendered images in the PartNet   \citep{mo2019partnet} dataset. Additionally, we collect HOI images corresponding to the affordance types from IAGNet   \citep{yang2023grounding} to alleviate GPT's hallucination issues, enabling better understanding. Fig. \ref{fig1} illustrates our ways of generating instructions in detail. 

% \begin{figure*}
%   \centering

%   \hfill
%   \begin{subfigure}{0.28\linewidth}
%     \fbox{\rule{0pt}{2in} \rule{.9\linewidth}{0pt}}
%     \caption{Another example of a subfigure.}
%     \label{fig:short-b}
%   \end{subfigure}
%   \caption{Example of a short caption, which should be centered.}
%   \label{fig:short}
% \end{figure*}

\subsection{Statistics and Analysis}

Our dataset comprises 162,386 instruction-point cloud pairs in the single affordance segmentation setting and 20,847 pairs in the sequential affordance segmentation setting, making up a total of 18,371 point cloud instances across 23 object categories. 

% As shown in Figure \ref{ciyun}, we have visualized word clouds for the object categories, affordance types, and the instructions in our dataset, demonstrating the richness of our dataset.

% \begin{figure}[htbp]
%     \centering
    
%     \begin{subfigure}{0.325\linewidth}
%         \centering
%         \includegraphics[width=\linewidth]{img/ciyun2.png}
%         \caption{Instructions}
%     \end{subfigure}
%     \hfill
%     \begin{subfigure}{0.325\linewidth}
%         \centering
%         \includegraphics[width=\linewidth]{img/ciyun1.png}
%         \caption{Affordance Types}
%     \end{subfigure}
%     \hfill
%     \begin{subfigure}{0.325\linewidth}
%         \centering
%         \includegraphics[width=\linewidth]{img/ciyun3.png}
%         \caption{Object Categories}
%     \end{subfigure}
    
%     \caption{Word clouds of (a) instructions, (b) affordance types, and (c) object categories.}
%     \label{ciyun}
% \end{figure}

Based on the complexity of the instructions, we have divided them into two settings. The first setting is based on instructions that can only infer a single specific affordance, which we define as the ``Single Affordance Segmentation Task”.  The second setting is based on instructions that 
can infer a combination of multiple affordances in sequence, which we define as the ``Sequential Affordance Segmentation Task”.

Inspired by the evaluation settings presented in LASO   \citep{li2024laso} and IAGNet   \citep{yang2023grounding}, we propose two types of distinct dataset configurations including \textit{Seen} and \textit{Unseen}. \textit{Seen}: This default setting maintains similar distributions of object classes and affordance types across both training and testing phases. \textit{Unseen}: This configuration is specifically designed to evaluate the model's ability to generalize affordance knowledge. In this setup, certain affordance-object pairings are deliberately omitted from the training set but introduced during testing. For instance, while the model may learn to grasp objects like bags and mugs during training, it is expected to generalize the “grasp” affordance to earphones, a pairing absent from the training data.

\section{Method}

\subsection{Architecture Overview}
\begin{figure*}[h]
\centering
\includegraphics[width=1\linewidth]{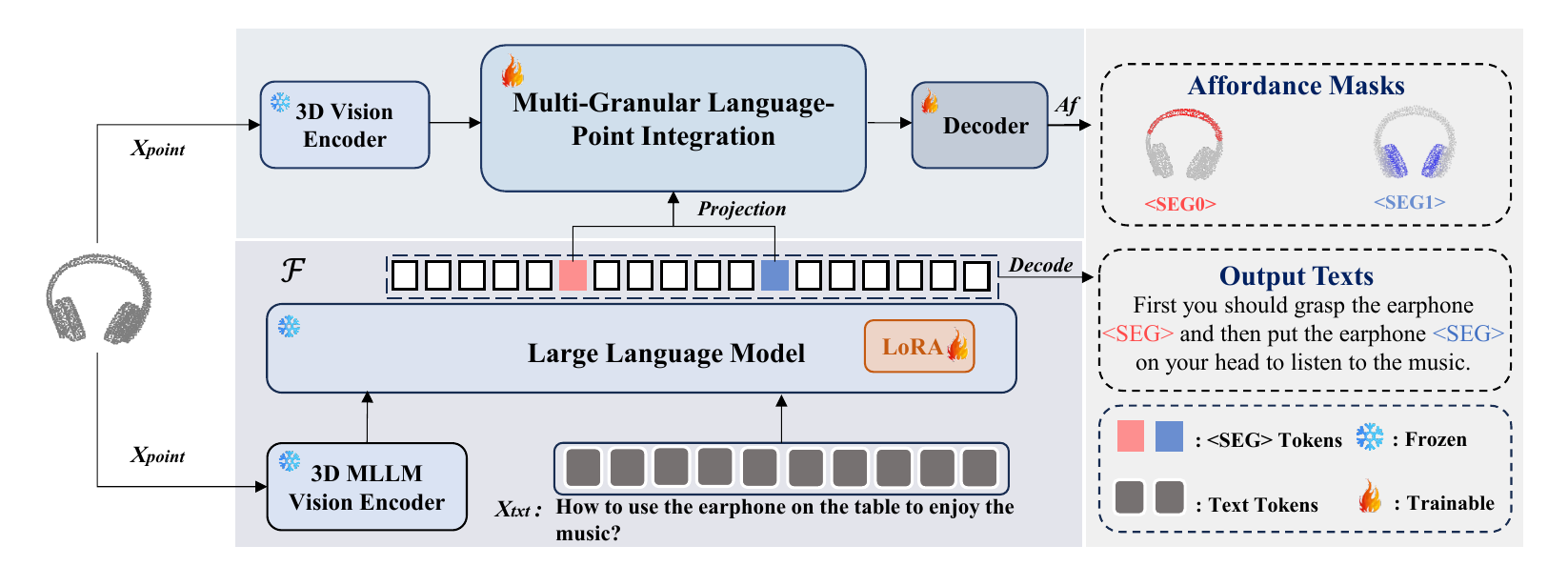}
\caption{\textbf{Main Pipeline.} Given the point clouds of the target objects and a piece of complex human instruction, SeqAfford first reasons from this instruction and decomposes it into several hidden \texttt{<SEG>} tokens extracted from the last-layer embeddings, each representing an intermediate affordance segmentation result. Then, for each \texttt{<SEG>}, the point features extracted by the 3D vision encoder dynamically interact with the \texttt{<SEG>} token before being sent to the decoder for mask generation. The interaction is achieved through multi-granular language-point integration, synergizing both reasoning and affordance segmentation. We use LoRA for efficient fine-tuning.}

\label{fig2}
\end{figure*}

The overall architecture of SeqAfford is presented in Fig. \ref{fig2}. Generally, SeqAfford mainly consists of three components: 1) a 3D vision encoder benefited from large-scale 3D representation learning, which provides solid foundations for dense prediction tasks; 2) a 3D Multi-modal Large Language Model (MLLM) $\mathcal{F}$ that exhibits affordance reasoning ability with the aid of internalized world knowledge; 3) a Multi-Granular Language-Point Integration module that considers the effective integration the point features and the segmentation tokens of MLLM, synergizing both reasoning and segmentation tasks from a multi-granular feature perspective.

%3) Following PointNet++   \citep{qi2017pointnet++}, a Feature Propagation Module upsamples and propagates the sparse, semantically rich point cloud features extracted by 3D Encoder, resulting in dense point cloud features that are rich in semantics; 

\subsection{Network Architecture}

\textbf{3D MLLM Backbone.} Recently, a series of 3D multi-modal language models have been proposed to deepen the understanding of open-world 3D objects, among which ShapeLLM is recently pretrained for understanding various embodied interactions. In light of this, we adopt ShapeLLM   \citep{qi2024shapellm} as our backbone, denoted as $\mathcal{F}$, where the point cloud encoder ReCon++ is pre-trained via multi-view distillation based on ReCon  \cite{qi2023contrast}, and the LLM is drawn from LLaMa  \cite{touvron2023llama}. Previous work on 3D affordance tasks typically employed 3D backbones  \cite{yang2023grounding, nguyen2023open} or utilized separate point-language encoders  \cite{li2024laso}, which may fall short in reasoning and open-world generalization abilities. Here, we leap ahead by directly utilizing a unified 3D MLLM instead of relying solely on pure LLMs or other visual structures as our backbones, for two main reasons: 1) it opens new possibilities for open-world 3D objects understanding, bolstering the generalization of unseen objects or affordances; 2) it internalizes affordance perception ability, compressing it into natural language form, thus preparing for the subsequent affordance reasoning task. %with its point cloud encoder ReCon++}, pre-trained on a large-scale point cloud-text paired dataset. The text tokenizer is the same one used in LLaVA   \citep{liu2024visual}.  

\noindent \textbf{Squential Affordance Reasoning.} Despite the efficacy of 3D MLLMs in aligning 3D representations with natural language, they are primarily designed for object-oriented text generation tasks and lack the capability for 3D dense prediction tasks, particularly in fine-grained affordance segmentation.
To encapsulate the segmentation ability into 3D MLLMs, a specific segmentation token \texttt{<SEG>} can be appended to the vocabulary set of the MLLM, inspired by  \cite{lai2024lisa}. 
%To further enable sequential reasoning of mulit-objects, we newly define \texttt{<SEG\_0>} ... \texttt{<SEG\_K>} in the vocabulary set, where $K$ is set larger than the possible maximum length of affordance sequences.

Formally, when provided with point cloud ${\bf X}_{\text{point}}$ and a text instruction  ${\bf X}_{\text{txt}}$ that demonstrates the user intentions on these potential objects, the 3D MLLM absorbs this multi-modal information and generates a text response $\tilde{\bf y}_\text{txt}$. It can be formulated as
\begin{equation}
    \tilde{\bf y}_\text{txt} = \mathcal{F}({\bf X}_{\text{point}}, {\bf X}_{\text{txt}}),
\end{equation}
where the output $\tilde{\bf y}_{txt}$ would include several \texttt{<SEG>} tokens, where a single \texttt{<SEG>} indicates a segmentation result within the sequence. We then extract the last-layer embeddings $\{{\bf h}^{(i)}_\text{seg}\}_{i=0}^{S-1}$ corresponding to the \texttt{<SEG>} tokens, where $S$ is the number of predicted affordance sequences. Afterwards, an MLP projection layer to obtain $\{{\bf H}^{(i)}_{seg}\}_{i=0}^{S-1}$ as follows
\begin{equation}
    {\bf H}^{(i)}_\text{seg} = \text{Proj}({\bf h}^{(i)}_\text{seg}).
\end{equation}

% \begin{figure}[h]
% \centering
% \includegraphics[width=1\textwidth]{ICLR 2025 Template/img/upsample.pdf}
% \caption{\textbf{Feature Propagation.}}
% \label{fig2}
% \end{figure}     

\noindent \textbf{Multi-Granular Language-Point Integration.} 
% \textbf{3D Encoder and Feature Propagation.} 
After obtaining several segmentation tokens that indicate a sequence of regions for reasoning where the given objects can be afforded, the remaining work entails integrating the abstracted reasoning results into 3D point clouds for dense affordance predictions. Therefore, the multi-granular language-point integration module mainly consists of two stages: 1) the multi-granular feature propagation process, which iteratively up-samples the point cloud features into dense features with multiple granularities considered; 2) the point-language integration stage, which distills the informative language features (i.e. the segmentation tokens) into the dense visual features (i.e. the 3D point features), and fuses the integrated dense features with global sparse features for final affordance segmentation. This serves as a crucial step towards reasoning and segmenting affordances in a cohesive framework.

\begin{figure*}[htbp]
\centering
\includegraphics[width=0.8\linewidth]{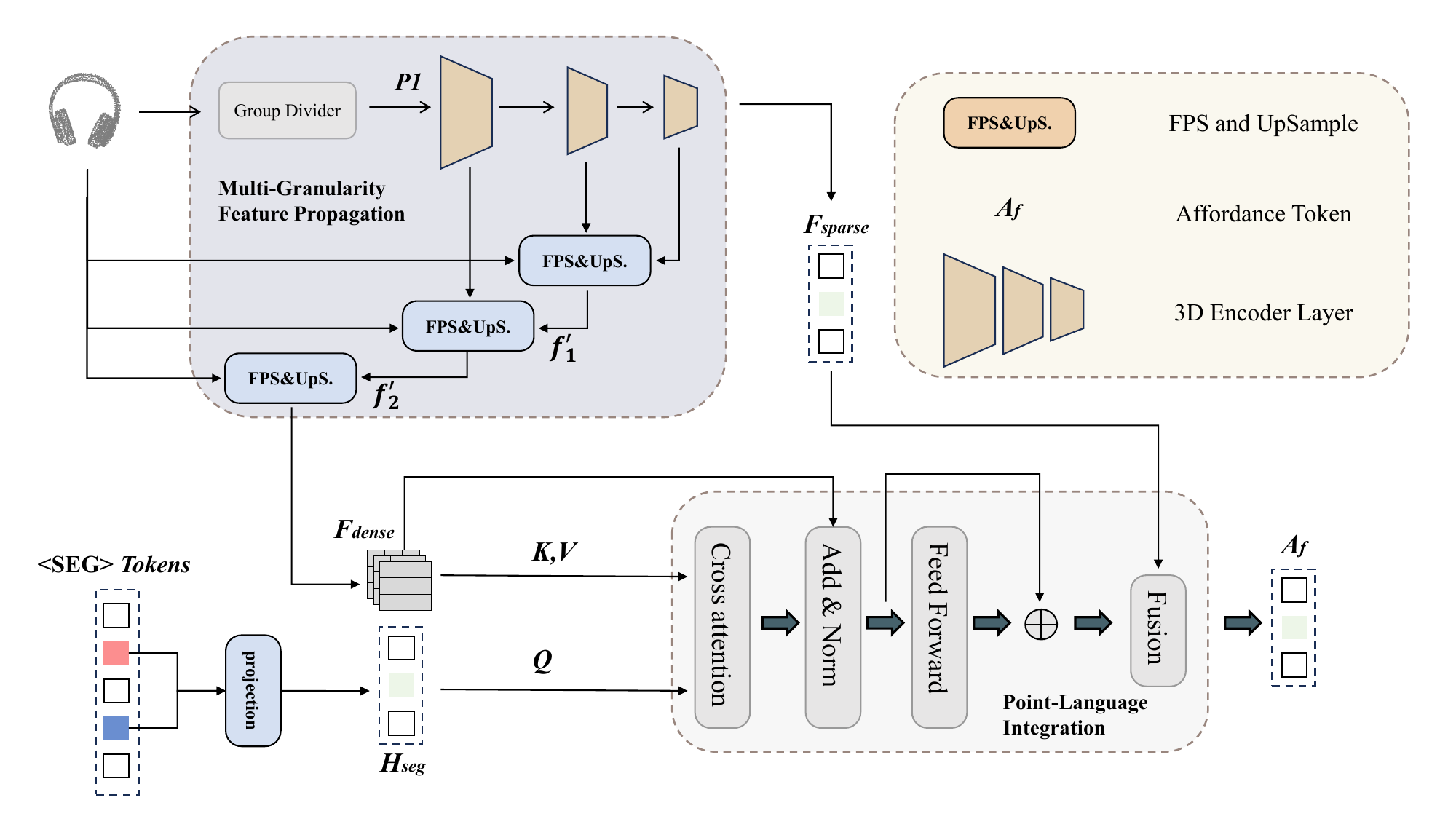}
\caption{\textbf{Multi-Granular Language-Point Integration Module.} We propose an interaction module between \texttt{<SEG>} tokens from LLM and point features from the 3D vision encoder, to synergize both reasoning and segmentation in a cohesive framework. This module consists of the multi-granular feature propagation process, and the point-language integration stage.}
\label{fig3}
\end{figure*}

To enable multi-granular feature propagation,  as shown in Fig. \ref{fig3}, we hierarchically up-sample (UpS. in Fig. \ref{fig3}) the intermediate features from the 3D encoder,  and propagate features through farthest point sampling (FPS in Fig. \ref{fig3}) process to sequentially generate ${\bf f}_1$, $ {\bf f}_2$, and the ultimate dense feature ${\bf f}_{\text{dense}}$. During the up-sampling process of intermediate features, various feature up-sample techniques are adopted, inspired by PointNet++   \citep{qi2017pointnet++} and DGCNN  \cite{wang2019dynamic}. More details about the multi-granular feature propagation process are revealed in the supplementary details. 

During the point-language integration stage, the model takes the dense point cloud features ${\bf f}_\text{dense}$, the sparse point cloud features  ${\bf f}_\text{sparse}$ and the instruction-rich $\bf H^{(i)}_\text{seg}$ as input. The ${\bf H}^{(i)}_\text{seg}$ and dense features ${\bf f}_\text{dense}$ are used as $\bf Q$ and $\bf K$, $\bf V$ respectively to perform cross-attention, and $\text{FFN}$, and the results obtained are fused with the sparse features ${\bf f}_\text{sparse}$ to get  ${\bf A}^{(i)}_f$. Finally, the decoder takes ${\bf A}^{(i)}_f$ as the input to get affordance mask $\tilde{\bf y}_\text{mask}$
\begin{align}
    {\bf A}^{(i)}_f &= \mathcal{G}({\bf f}_\text{dense},{\bf f}_\text{sparse}, {\bf H}^{(i)}_\text{seg}), \quad 
    \tilde{y}_\text{mask} = \mathcal{D}({\bf A}^{(i)}_f),
\end{align}

where $\mathcal{G}$ denotes point-language integration, and $\mathcal{D}$ denotes the decoder.

\subsection{Training objectives}

Our objective is to train an end-to-end MLLM capable of generating diverse texts while simultaneously predicting point-wise affordance masks. To this end, we employ auto-regressive cross-entropy loss $\mathcal{L}_c$ for text generation, Dice loss $\mathcal{L}_d$ and Binary Cross-Entropy loss $\mathcal{L}_\text{b}$ for guiding the segmentation mask prediction. 

\begin{equation}
\begin{aligned} 
 \mathcal{L} = \lambda_{c}  \mathcal{L}_{c}({\bf y}_\text{txt}, \tilde{\bf y}_\text{txt}) + \lambda_{b}  \mathcal{L}_{b}({\bf y}_\text{mask}, \tilde{\bf y}_\text{mask}) + \\  \lambda_{d}  \mathcal{L}_{d}({\bf y}_\text{mask}, \tilde{\bf y}_\text{mask}),
 \end{aligned} 
\end{equation}

where the weights $\lambda_c,\lambda_b,\lambda_d$ are utilized to balance the different loss items.

\section{Experiment}

\begin{table}[ht]
\centering
\begin{tabular}{@{}lccccc@{}}

\toprule
& Method & \textit{mIoU}$\uparrow$ & \textit{AUC}$\uparrow$ & \textit{SIM}$\uparrow$ & \textit{MAE}$\downarrow$ \\
\midrule
\multirow{6}{*}{\rotatebox{90}{Seen}} 
& ReferTrans  \citep{li2021referring} & 11.4 & 77.2 & 0.449 & 0.135 \\
& ReLA  \citep{liu2023gres} & 12.1 & 76.3 & 0.480 & 0.130 \\
& 3D-SPS  \citep{luo20223d} & 10.1 & 75.2 &0.413 & 0.141 \\
& IAGNet  \citep{yang2023grounding} & 14.2 & 81.7 & 0.510 & 0.117 \\
& PointRefer  \citep{li2024laso} & 16.3 & 84.3 & 0.568 & 0.108 \\
\rowcolor{gray!20}
& Ours & \textbf{19.5} & \textbf{86.9} & \textbf{0.594} & \textbf{0.098} \\
\midrule
\midrule

\multirow{6}{*}{\rotatebox{90}{Unseen}} 
& ReferTrans  \citep{li2021referring} & 9.1 & 67.4 & 0.427 & 0.151 \\
& ReLA  \citep{liu2023gres} & 9.3 & 68.2 & 0.423 & 0.147 \\
& 3D-SPS  \citep{luo20223d} & 7.1 & 66.9 &0.397 & 0.162 \\
& IAGNet  \citep{yang2023grounding} & 11.7 & 73.6 & 0.438 & 0.143 \\
& PointRefer  \citep{li2024laso} & 12.4 & 76.1 & 0.502 & 0.132 \\
\rowcolor{gray!20}
& Ours & \textbf{13.8} & \textbf{82.4} & \textbf{0.518} & \textbf{0.128} \\

\midrule
\midrule

\multirow{6}{*}{\rotatebox{90}{Squential}} 
& ReferTrans*   \citep{li2021referring} &10.8  &74.5  &0.425  &0.142 \\
& ReLA*   \citep{liu2023gres} &11.4  &74.8  &0.463 &0.136 \\
& 3D-SPS*   \citep{luo20223d} &9.9  & 73.1 &0.407 & 0.148 \\
& IAGNet*   \citep{yang2023grounding} & 13.5 &78.2  & 0.496 & 0.131 \\
& PointRefer*   \citep{li2024laso} & 14.3 &80.7 & 0.521 & 0.124 \\
\rowcolor{gray!20}
& Ours &\textbf{14.6}  &\textbf{84.2}  & \textbf{0.573} &\textbf{0.118}  \\
\bottomrule
\end{tabular}
\caption{\textbf{Main Results.} The overall results of all comparative methods, the best results are in bold. Seen and Unseen are two partitions of the Single Affordance segmentation dataset. AUC and mIoU are shown in percentage. * means that baseline methods use \textbf{ground-truth} sequential order as all existing methods cannot predict sequential affordances.}
\label{result1}
\end{table}

We conduct extensive experiments to evaluate the effectiveness of our proposed dataset, task, and method, including both Single and Sequential Affordance segmentation tasks. In Sec. \ref{funda}, we assess the capability of our model to ground the Single Affordance with simple instruction. Sec. \ref{squen} studies a more challenging task where the model is requested to predict the sequential affordances. Various ablation experiments on our model are performed in Sec. \ref{abb}.

\noindent \textbf{Implementation Details.} We employ ShapeLLM   \citep{qi2024shapellm} as our 3D MLLM module in this paper, with the ShapeLLM-7B checkpoint as the default setting and we freeze the 3D encoder during training. We adopt Uni3D as the 3D vision encoder to enhance the 3D dense prediction tasks. Unless otherwise stated, the projection layer is implemented as a multi-layer perceptron. We employ LoRA   \citep{hu2021lora}  for efficient fine-tuning and set the rank of LoRA to 8 by default. Additionally, we utilize AdamW   \citep{loshchilov2017decoupled} optimizer with the learning rate and weight decay set to 0.0002 and 0,  respectively. We adopt a cosine learning rate scheduler, with the warm-up iteration ratio set to 0.03. All attentions in ShapeLLM \citep{qi2024shapellm} are replaced by flash-attention   \citep{dao2022flashattention} during training. The training is done on one A100 GPU for 10 epochs for the main experiments and during training, we use all mentioned datasets in Sec. \ref{data} for joint training by leveraging task-specific prompts. For evaluation on a specific dataset, we finetune the trained model on the corresponding dataset.

\noindent \textbf{Evaluation Metrics and Baseline.} To provide a comprehensive and effective evaluation, we follow previous works and finally chose four evaluation metrics: Area Under the Curve (AUC)   \citep{lobo2008auc}, Mean Intersection Over Union (mIOU)   \citep{rahman2016optimizing}, SIMilarity (SIM)   \citep{swain1991color} and Mean Absolute Error (MAE)   \citep{willmott2005advantages}.To the best of our knowledge, LASO   \citep{li2024laso} is the most similar to our work. For a thorough comparison of our method, we conduct comparisons based on its setup in the Single Affordance segmentation task, comparisons with other baselines were also implemented following the approach mentioned in it. While in the sequential affordance segmentation task, we offer sequential information to these models enabling them to perform ``sequential`` reasoning.

\begin{figure*}[h]
\centering
\includegraphics[width=0.85\textwidth]{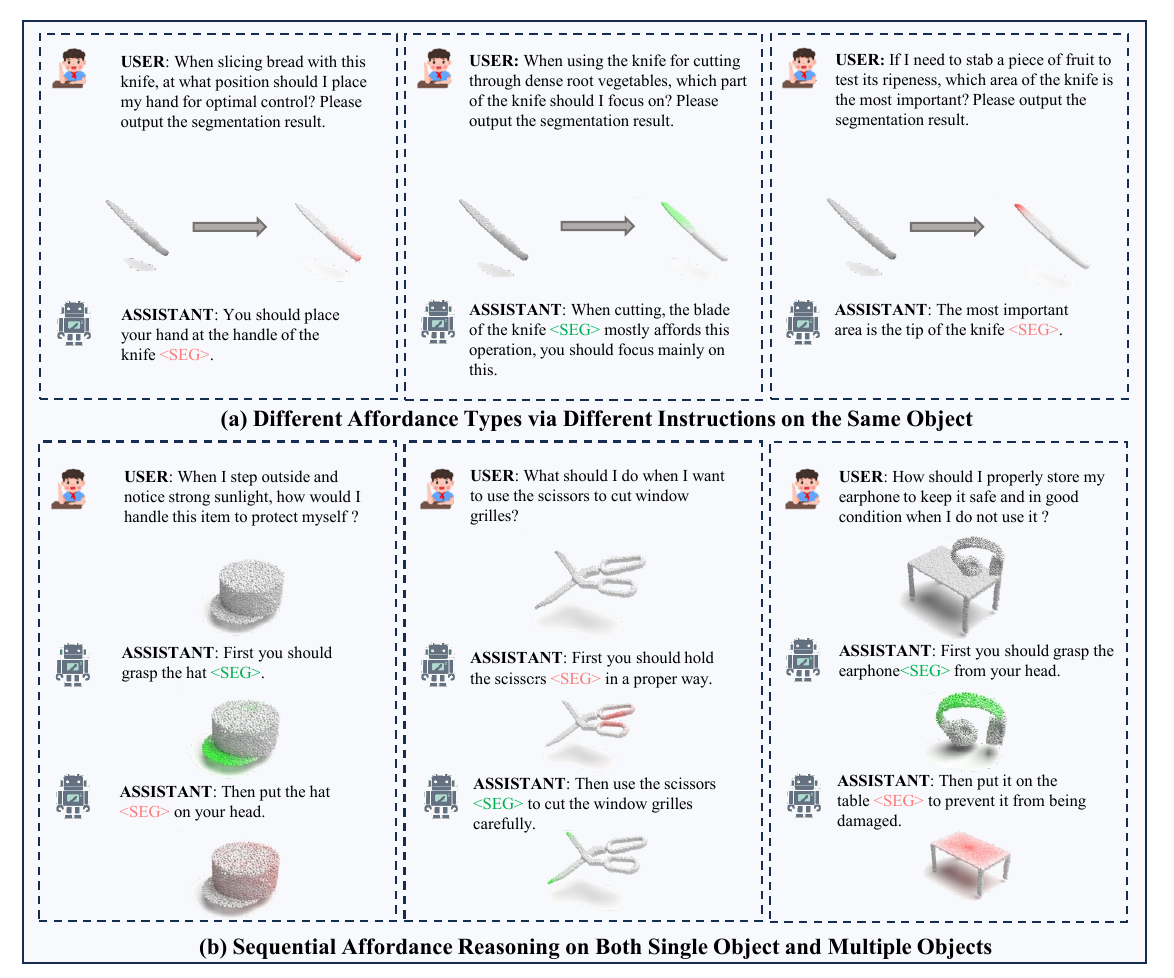}
\caption{Qualitative results of our model. SeqAfford understands human instruction and accurately segments the target affordance.}
\label{vis}
\end{figure*}    

\subsection{Results on Language-Guided Single Affordance Segmentation Task}
\label{funda}
As detailed in Table. \ref{result1}, our model demonstrates superior performance across all evaluation metrics compared to the baseline methods.
Unlike conventional segmentation tasks, language-guided Single Affordance segmentation demands not just identification but the integration of perception and cognition, necessitating the model's reasoning capabilities and access to world knowledge. Existing approaches struggle with implicit queries due to their lack of integration of perception and cognition, which further underscores the task's inherent challenges. In contrast, our model leverages MLLMs to bridge this gap, demonstrating superior performance by comprehending and interpreting the queries accurately.

\subsection{Results on Language-Guided Sequential Affordance Reasoning Task}
\label{squen}
The sequential affordance reasoning task implies the ability to infer multiple affordances from a single text, which requires a more profound integration of cognitive and perceptual capabilities compared to the Single Affordance segmentation task. In our model, to make the Multimodal Large Language Model (MLLM) more comprehensible, we use the format \texttt{<SEG>} to represent the sequence of affordances and decode to obtain the affordance mask based on them. The previous models did not possess this capability. To compare them with our method, we used GPT to decompose the original instructions into new sequence instructions, then fed the decomposed instructions separately as inputs to these models, enabling them to perform ``sequential” reasoning. The main results are shown in Table. \ref{result1}, and we speculate that the reason our model performs better is that, compared to LASO   \citep{li2024laso} which merely uses language models to encode the input text, we have introduced a multimodal 3D large language model. This model has a much stronger capability for integrating 3D data and text than a purely linguistic model, and it possesses a richer knowledge of the world, enabling it to handle multimodal sequential tasks more effectively. The sequential reasoning results are visualized in Fig. \ref{vis} (b), where our model can understand how a task is connected with sequential actionable affordances involving multi-objects. This ability is not only attributed to the challenging benchmark collected from diverse sources but also the powerful world knowledge internalized in 3D MLLMs.

\subsection{Ablation Study}
\label{abb}

We conduct various ablation studies to assess the impact of different model implementations on our model SeqAfford's performance, including the multi-granular language-point integration module and the different choices of 3D  vision encoder backbones.

\noindent \textbf{Multi-Granular Language-Point Integration Module}. Introducing the Multi-Granular Language-Point Integration (MGLP)  module results in a substantial improvement over the baseline, as shown in Table. \ref{spf}. This underscores our method's capability to minimize information loss and this indicates that our method enables deep integration of high-dimensional semantic features representing instructions with dense features from point clouds and semantically rich but sparse features from point clouds, thereby making affordance reasoning more effective.

\noindent \textbf{Choice of 3D  Vision Encoder Backbone}. We conducted ablation experiments to study the influence of the 3D  Vision Encoder backbone, and we investigated the performance of 3D SeqAfford with some alternative backbones. As shown in Table. \ref{3Dbackbone}, Uni3D   \citep{zhang2023uni3d} performs better in this task due to its strong representation ability, Thus, we set it as our default 3D vision encoder backbone.

\begin{table}[ht]
\centering
\scalebox{0.9}{
\begin{tabular}{@{}lcccccc@{}}
\toprule
& Variants &Task & \textit{mIoU}$\uparrow $& \textit{AUC}$\uparrow $& \textit{SIM}$\uparrow$ & \textit{MAE}$\downarrow$ \\
\midrule
&w/o MGLP  &Single&12.1 & 83.4 &0.552  &0.117  \\
\rowcolor{gray!20}
& Ours &Single&\textbf{19.5}  &\textbf{86.9}  & \textbf{0.594} &\textbf{0.098}  \\
&w/o MGLP  &Sequential&11.7 &80.3  &0.518  &0.129  \\
\rowcolor{gray!20}
& Ours &Sequential&\textbf{14.6}  &\textbf{84.2}  & \textbf{0.573} &\textbf{0.118}  \\

\bottomrule
\end{tabular}}
\caption{Ablation study on multi-granular language-point module.}
\label{spf}
\end{table}

\begin{table}[ht]
\centering
\scalebox{0.9}{
\begin{tabular}{@{}lcccccc@{}}
\toprule
& 3D Vision Backbone  & \textit{mIoU}$\uparrow$ & \textit{AUC}$\uparrow$ & \textit{SIM}$\uparrow$ & \textit{MAE}$\downarrow$ \\
\midrule
& ULIP   \citep{xue2023ulip} &17.9  &84.8  & 0.574 &0.109  \\
& OpenShape   \citep{liu2024openshape} &18.4  &85.3  & 0.582 &0.103  \\
& Recon++   \citep{qi2024shapellm} &19.1  &86.4  &0.588 &0.099  \\
\rowcolor{gray!20}
& Uni3D   \citep{zhang2023uni3d} &\textbf{19.5}  &\textbf{86.9}  & \textbf{0.594} &\textbf{0.098}  \\
\bottomrule
\end{tabular}}
\caption{Ablation study on the choices of 3D vision backbone.}
\label{3Dbackbone}
\end{table}

\section{Conclusion}
% In this paper, we introduce a novel task called the Sequential 3D Affordance Reasoning Task, designed to address complex user intentions that may involve sequential or long-horizon subtasks, transcending the conventional single-object, single-affordance paradigm. Our contributions include not only the construction of a large-scale instruction-tuning 3D affordance benchmark, comprising 180K instruction-point pairs collected from diverse sources, but also the introduction of an advanced 3D multimodal LLM capable of applying world knowledge to actively interpret complex user intentions and produce sequential, actionable affordance maps, accompanied by reasonable explanations. To conclude, SeqAfford propels 3D affordance fields forward by enabling deeper integration with LLM advancements, thus enhancing the model’s capacity to understand and respond to complex, sequential user intentions in real-world contexts. For future work, we intend to extend our research to encompass scene-level fine-grained reasoning, enabling more sophisticated and context-aware affordance segmentation that aligns with the complexities faced by embodied intelligent agents in real-world scenarios. 

In this paper, we introduce a novel task called the Sequential 3D Affordance Reasoning Task, designed to address complex user intentions that may involve sequential or long-horizon subtasks, transcending the conventional single-object, single-affordance paradigm. Our contributions include the construction of a large-scale 3D affordance benchmark comprising 180K instruction-point pairs collected from diverse sources, and the development of an advanced 3D multimodal LLM that leverages world knowledge to interpret complex user intentions, producing sequential, actionable affordance maps with reasonable explanations. SeqAfford advances 3D affordance reasoning by integrating LLM capabilities, enhancing the model’s ability to handle complex, sequential tasks in real-world contexts. Our future work will focus on scene-level fine-grained reasoning to enable more sophisticated and context-aware affordance segmentation for embodied intelligent agents.

%While our work represents a significant advancement in 3D affordance segmentation, there are still some limitations. One primary constraint is the reliance on high-quality 3D point cloud data, which may not always be readily available or easy to obtain in real-world settings. Additionally, our research primarily focuses on object-centric reasoning and lacks scene-level fine-grained reasoning, which is crucial for embodied intelligent agents. 

% For future work, we plan to address these limitations through several approaches. First, we aim to develop methods for augmenting or synthetically generating diverse 3D point cloud data to improve the robustness of our models in varied environments. Additionally, we intend to extend our research to encompass scene-level fine-grained reasoning, enabling more sophisticated and context-aware affordance segmentation that aligns with the complexities faced by embodied intelligent agents in real-world scenarios. 

\section{Acknowledgement }
This work was supported by the Shanghai Local College Capacity Building Program (23010503100), National Natural Science Foundation of China (No.62406195), MoE Key Laboratory of Intelligent Perception and Human-Machine Collaboration (ShanghaiTech University), HPC Platform of ShanghaiTech University and Shanghai Engineering Research Center of Intelligent Vision and Imaging.

{
    \small
    \bibliographystyle{ieeenat_fullname}
    \bibliography{main}
}

% WARNING: do not forget to delete the supplementary pages from your submission 

\clearpage
\setcounter{page}{1}
% \maketitlesupplementary
\onecolumn
\begin{center}
{ \linespread{1.5} \selectfont
\textbf{\Large SeqAfford: Sequential 3D Affordance Reasoning via \\
Multimodal Large Language Model} \\
}
\Large Supplementary Materials
\end{center}
\appendix
\setcounter{table}{0}   %从0开始编号，显示出来表会A1开始编号
\setcounter{figure}{0}

%定义编号格式，在数字序号前加字符“A"
\renewcommand{\thetable}{A\arabic{table}}
\renewcommand{\thefigure}{A\arabic{figure}}
\makeatletter

\startcontents[appendix]
\printcontents[appendix]{l}{1}{\section*{Supplementary organization:}\setcounter{tocdepth}{2}}

\section{Code and Dataset}
We will release our code and dataset once the paper is accepted.
\section{Textual Prompt Format}

This section outlines the specific design of the Textual Prompt component included in the four generative paradigms demonstrated earlier. By designing the Textual Prompt, it helps to enhance GPT-4's understanding and generalization ability for the relevant tasks, providing suitable introductions for subsequent tasks. We have utilized the principles of prompt engineering to design modules such as Role, Task, Example, and Instruction for the Textual prompt. Additionally, we have formatted the input in JSON format to improve GPT-4's comprehension of the text, standardizing the input of Object Name and Affordance Type.We show the details in Fig. \ref{tpf}.

\subsection{Role}
In the role module, we have preset the textual Prompt template as follows: "Role: You are an analytical assistant specializing in robotic affordance grounding. Your expertise is in creating questions that facilitate the training of robotic affordance grounding, enabling robots to reason about task execution, such as determining the appropriate part of an object to grasp."
\subsection{Task}
In the task module, we have preset the textual Prompt template as follows: "Task: Description: You will be provided with the name of an object. The robot is expected to use this tool to perform a variety of everyday tasks. Along with the tool name, you will receive an affordance type that the object can afford and a list of questions that have already been generated for this tool. Guidelines: Diversity: Aim for a wide range of tasks, ensuring that there is no overlap with previous ones. Daily Tasks: questions should be common and representative of the ones encountered in daily life. Leakage Avoidance: Ensure that the generated questions tasks can only afford the given affordance."

\subsection{Examples}
Examples illustrate the types of questions that can be generated based on the object name and affordance type. These examples guide the creator of the prompt in formulating new and diverse questions.
In the task module, we have preset the textual Prompt template as follows: "Examples: When cutting a rope with these scissors, which part of it should your palm touch? If you want to lift the bag, at which point is your finger most likely to carry it? Point out the areas on the microwave ideal for opening. How would you grasp the hat to best maintain its condition? If you want to boil water, at which points on the tap would you open the water valve? And so on...". 

\subsection{Instruction}
In the task module, we have preset the textual Prompt template as follows: "Instruction: With the provided OBJECT NAME: '+object+' and AFFORDANCE TYPE: '+affordance types+', generate fifteen new affordance grounding question tasks. Use the HISTORY of generated tasks as a reference to ensure compliance with the diversity guideline. Output should be in the JSON format."

\subsection{Template}
Here is the template that encapsulates all the elements of the textual prompt:
\begin{figure*}
    \centering
    \includegraphics[width=0.8\linewidth]{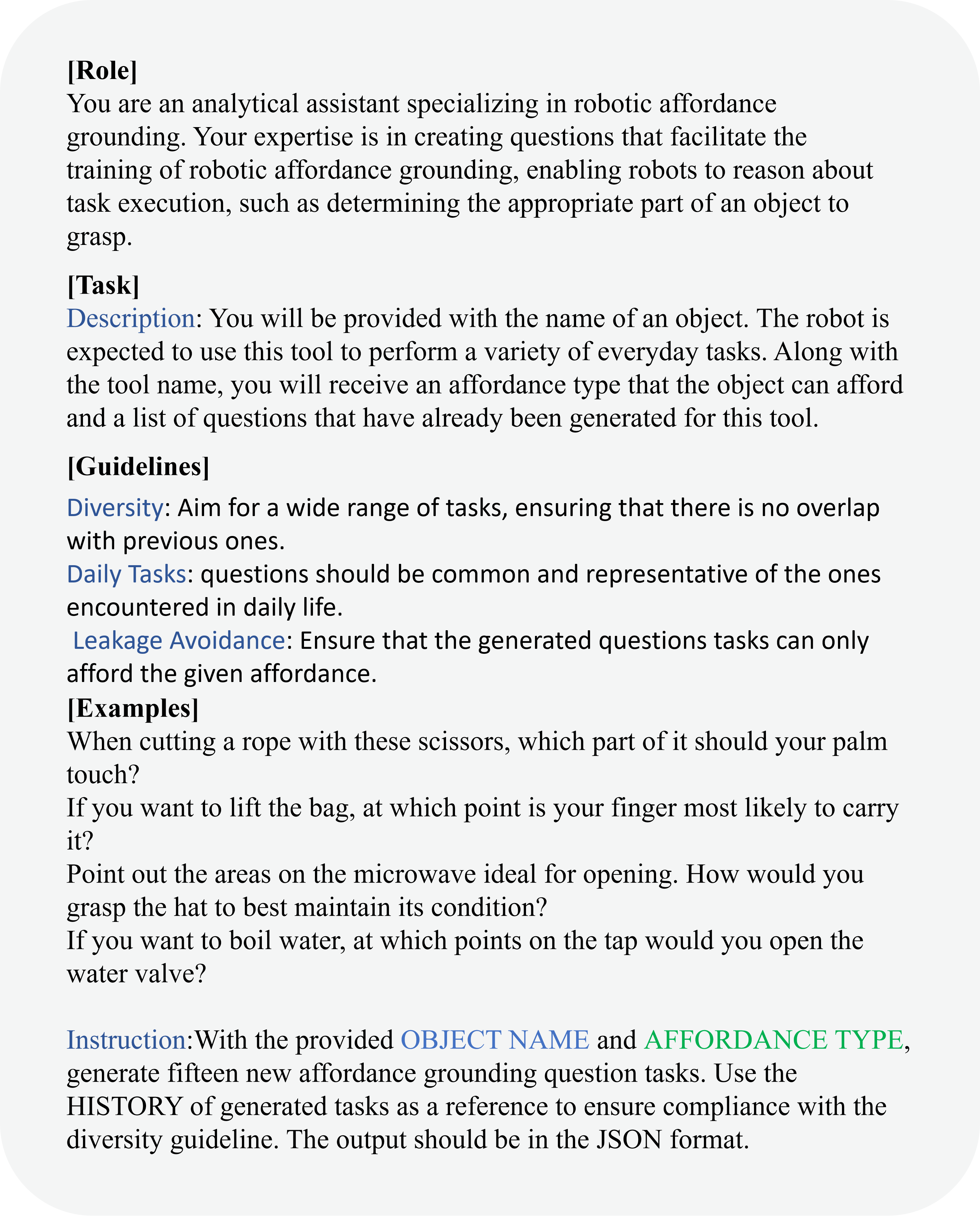}
    \caption{textual Prompt Format}
    \label{tpf}
\end{figure*}

\subsection{Genereted Instructions Example}
\begin{figure*}[htbp]
    \centering
    \includegraphics[width=1\linewidth]{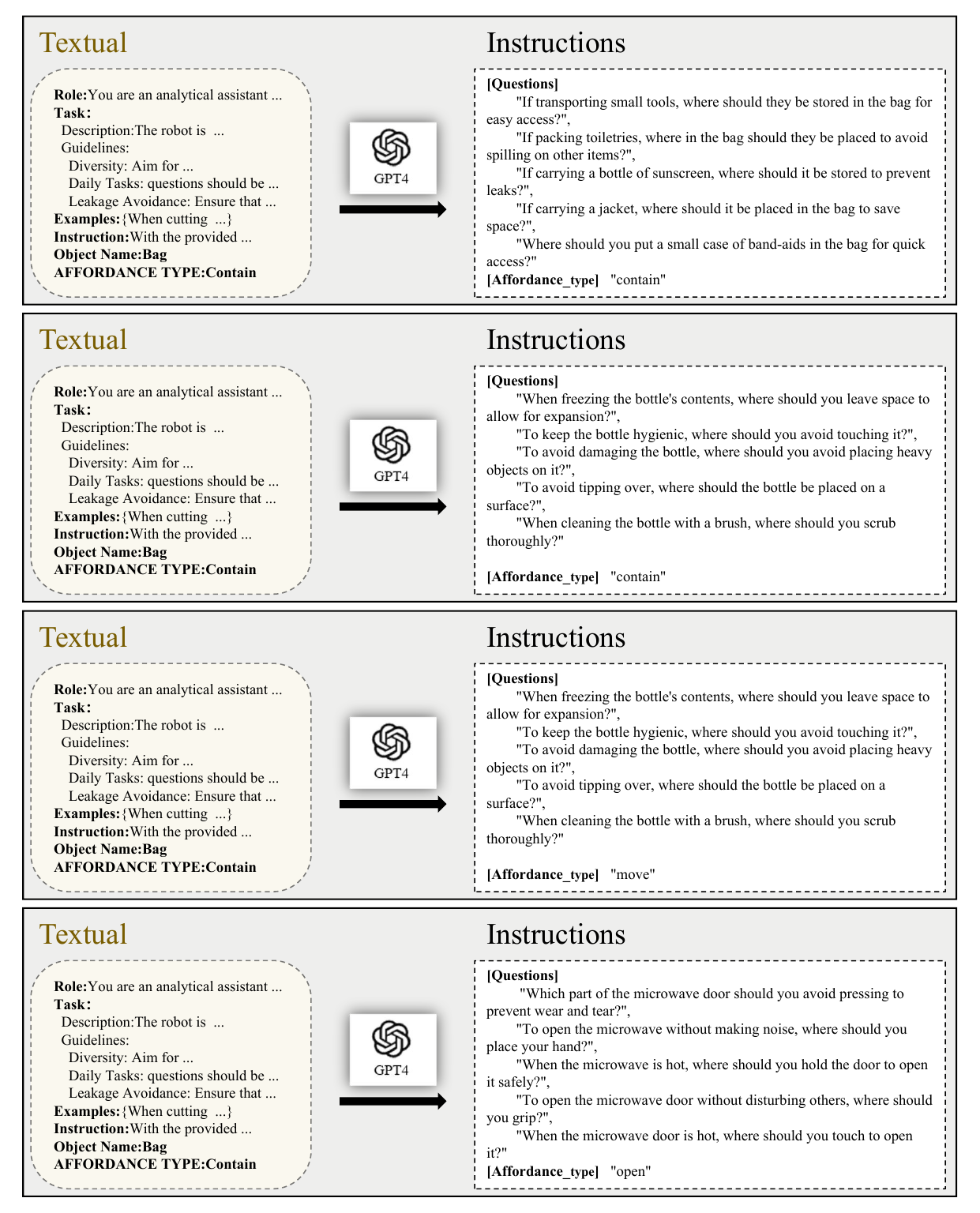}
    \caption{Genereted Instructions Example:Textual Prompt}
    \label{gie}
\end{figure*}

\begin{figure*}[htbp]
    \centering
    \includegraphics[width=1\linewidth]{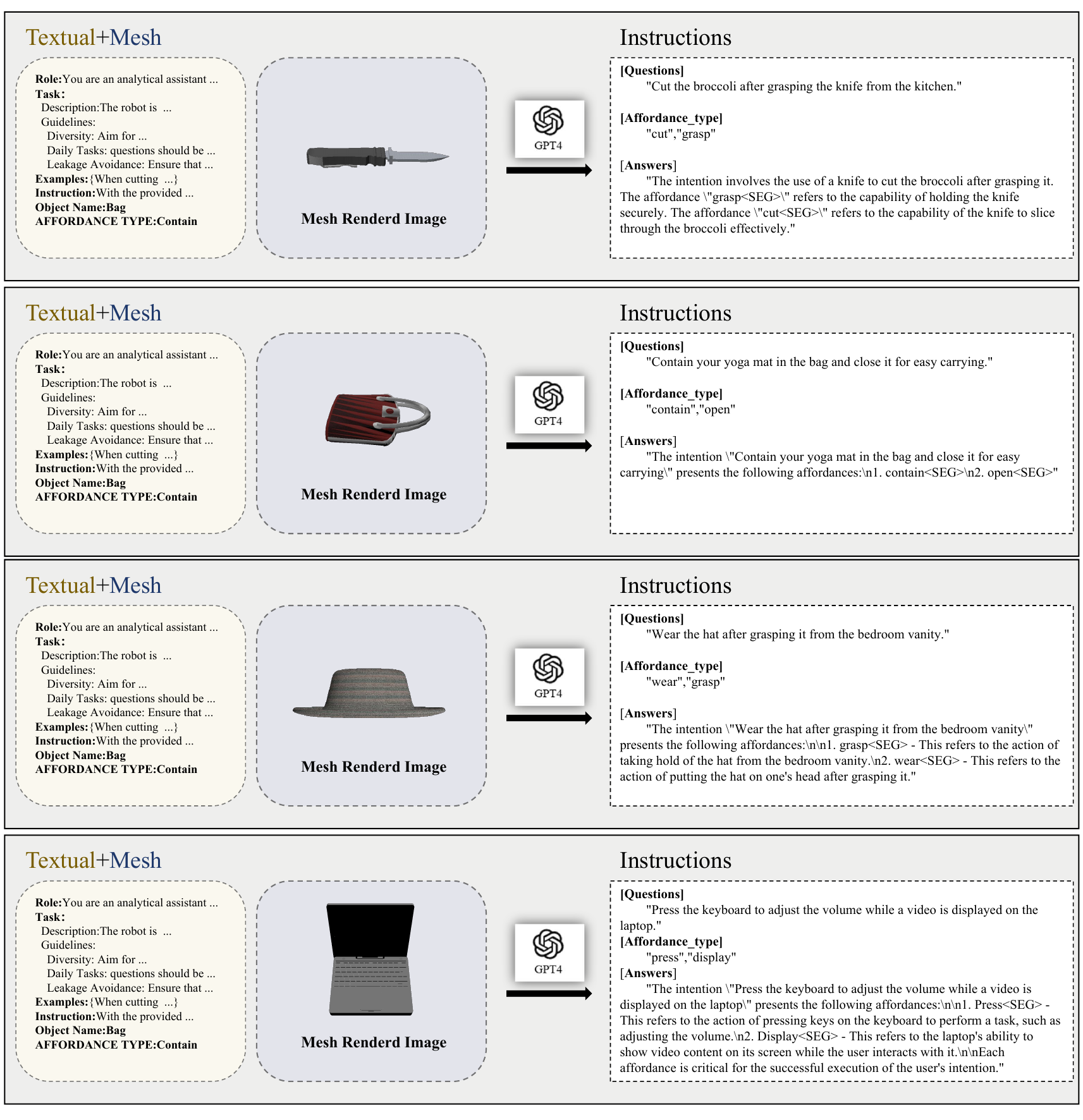}
    \caption{Genereted Instructions Example:Textual Prompt with Mesh Images}
    \label{tpm}
\end{figure*}

\begin{figure*}[htbp]
    \centering
    \includegraphics[width=1\linewidth]{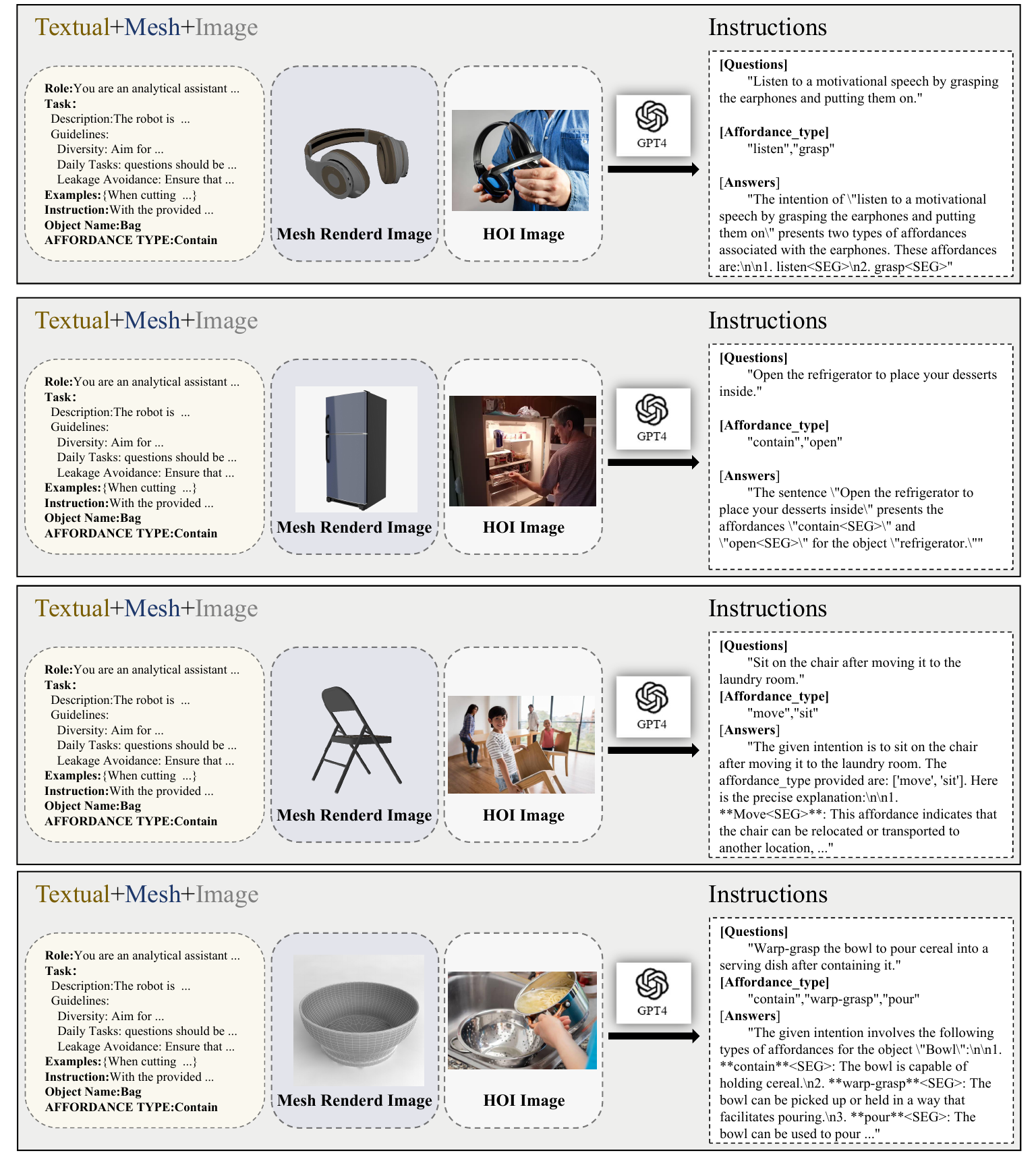}
    \caption{Genereted Instructions Example:Textual Prompt with Mesh and HOI Images}
    \label{hoi}
\end{figure*}

\begin{figure*}[htbp]
    \centering
    \includegraphics[width=1\linewidth]{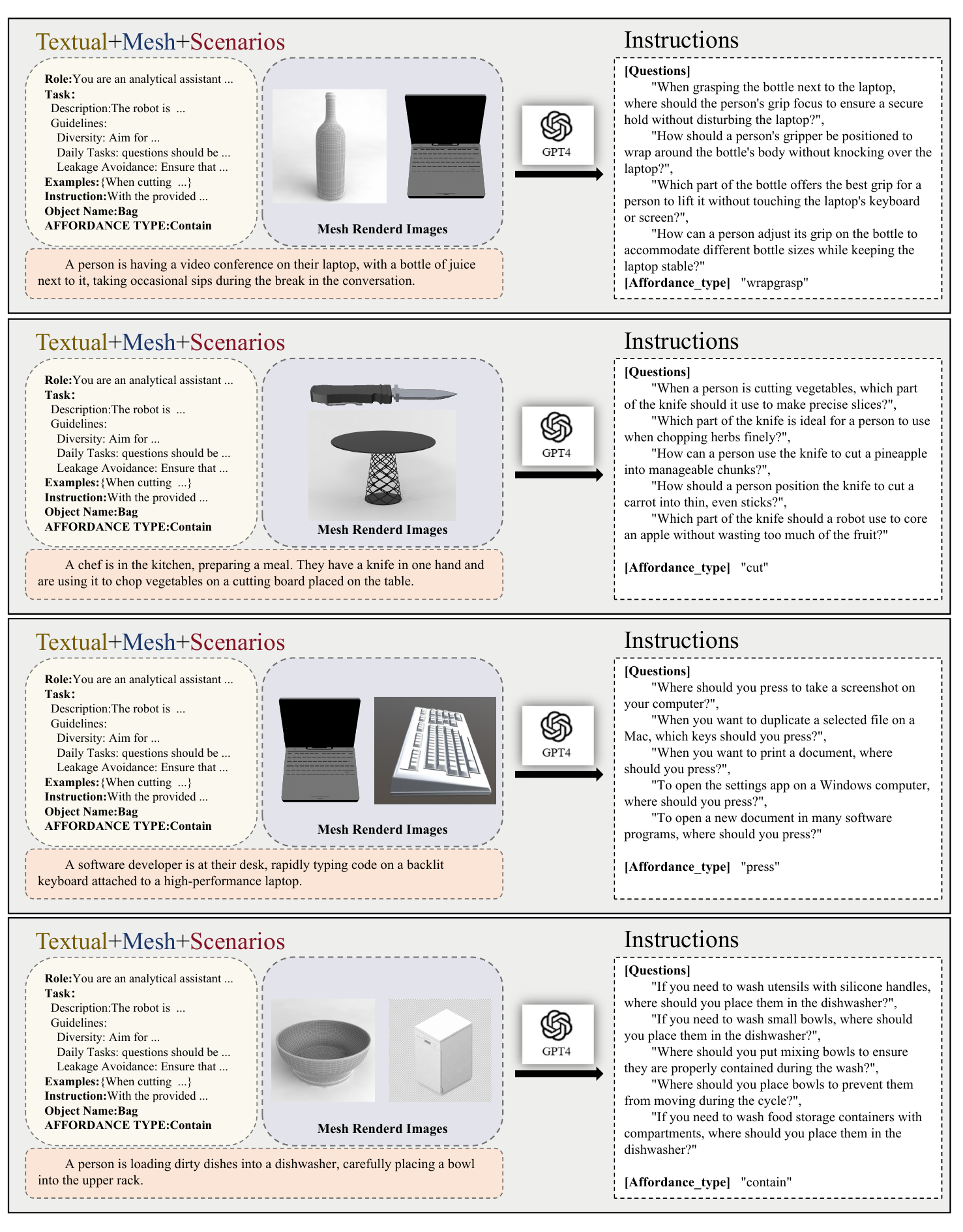}
    \caption{Genereted Instructions Example:Textual Prompt with Mesh Images and Scenarios Descriptions}
    \label{sce}
\end{figure*}

We illustrated detailed examples of our generated instructions in Fig. \ref{gie}, \ref{tpm}, \ref{hoi}, \ref{sce}


\section{Format of Questions and and Answers }
\begin{figure*}
    \centering
    \includegraphics[width=0.8\linewidth]{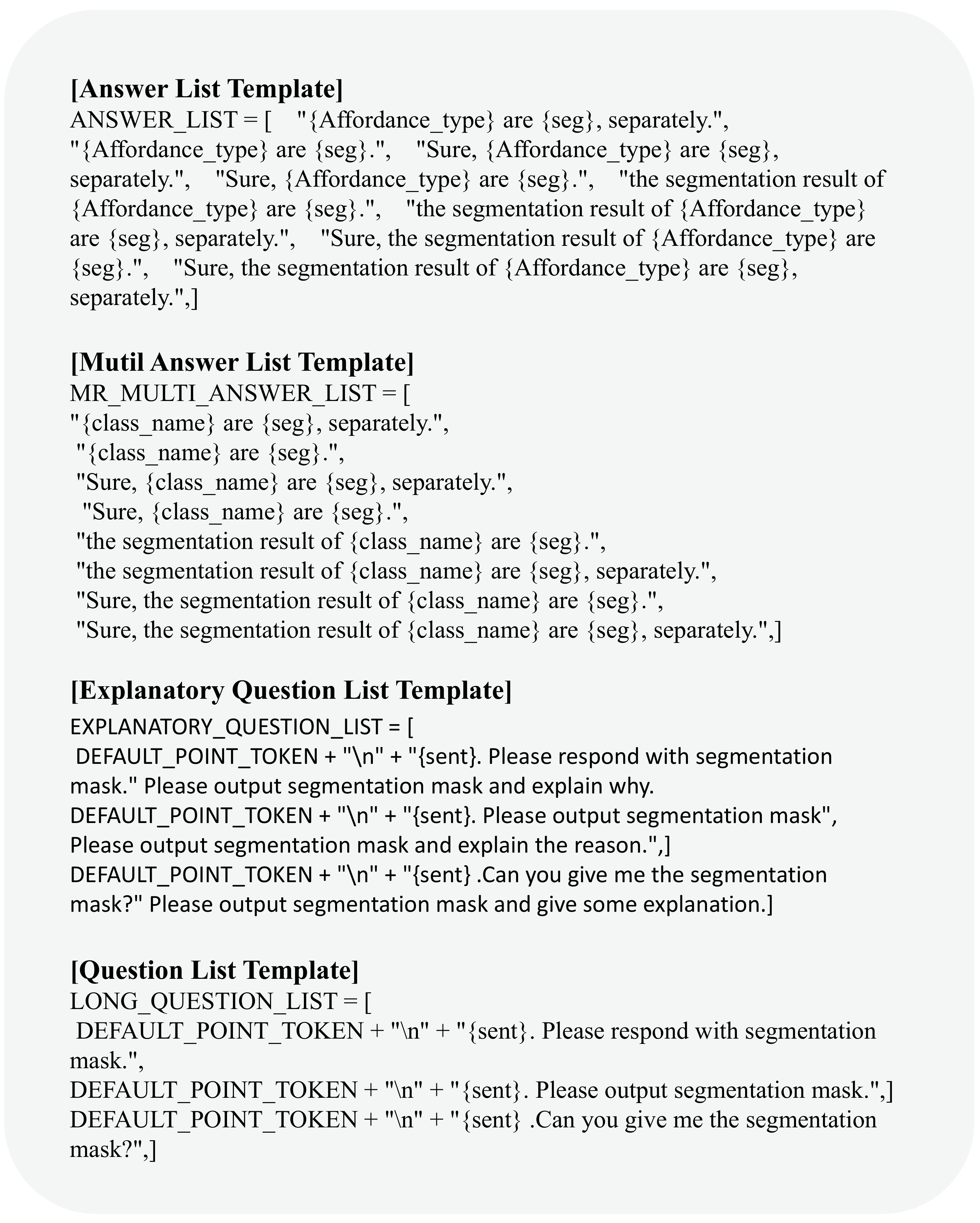}
    \caption{The details of our templates}
    \label{supfig5}
\end{figure*}

We have several formats of input, we show the Question List template , Explanatory Question List template and Answer List template in Fig. \ref{supfig5}

\begin{table*}
\caption{\textbf{Dataset Statistics}. This table showcases the distribution of our instruction-tuning benchmark for the Sequential Affordance Reasoning task, which introduces a wide range of variances and realistic simulations, including seen/unseen settings, single/sequential scenarios.}
\label{Table1}
\centering
\scalebox{0.9}{

\begin{tabular}{cccccc}
\toprule
\multirow{1}{*}{\bf Task} & \multirow{1}{*}{\bf Setting} & \multicolumn{2}{c}{\bf Train} & \multicolumn{2}{c}{\bf Test} \\

\multicolumn{1}{c}{} & \multicolumn{1}{c}{} & \multicolumn{1}{c}{\multirow{2}{*}{Shapes}} & \multicolumn{1}{c}{\multirow{2}{*}{Instruction-Point Cloud Pairs}} & \multicolumn{1}{c}{\multirow{2}{*}{Shapes}} & {\multirow{2}{*}{Instruction-Point Cloud Pairs}} \\
& & & & & \\
\midrule
Single & Seen & 16084 & 157,820 & 2287 & 4566 \\
Single & Unseen & 14307 & 124,140 & 2287 & 4566 \\
\midrule
Sequential & Seen & 8156 & 13393 & 2,946 & 7454 \\
\bottomrule
\end{tabular} }
\end{table*}
\section{Dataset Details}

\begin{figure*}
    \centering
    \includegraphics[width=0.8\linewidth]{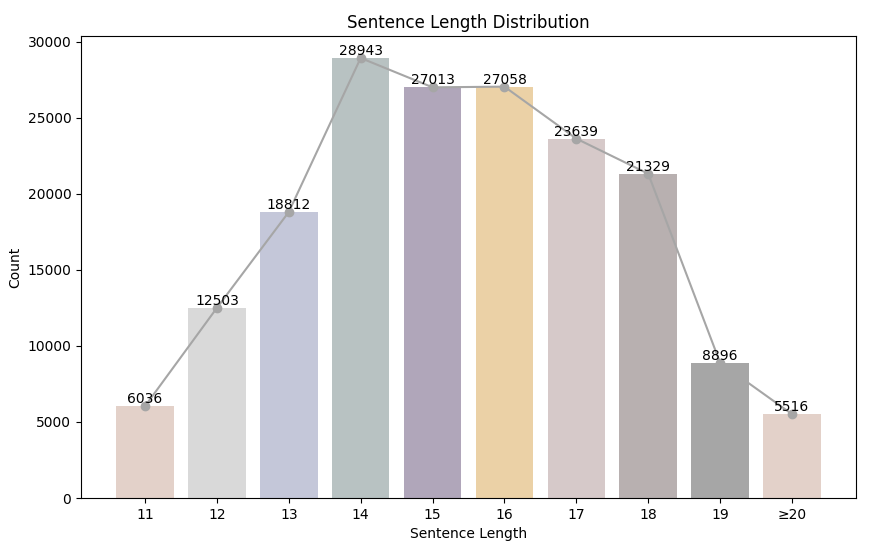}
    \caption{Question Length Distribution}
    \label{length-que}
\end{figure*}

\begin{figure*}[htbp]
    \centering
    
    \begin{subfigure}{0.325\textwidth}
        \centering
        \includegraphics[width=\textwidth]{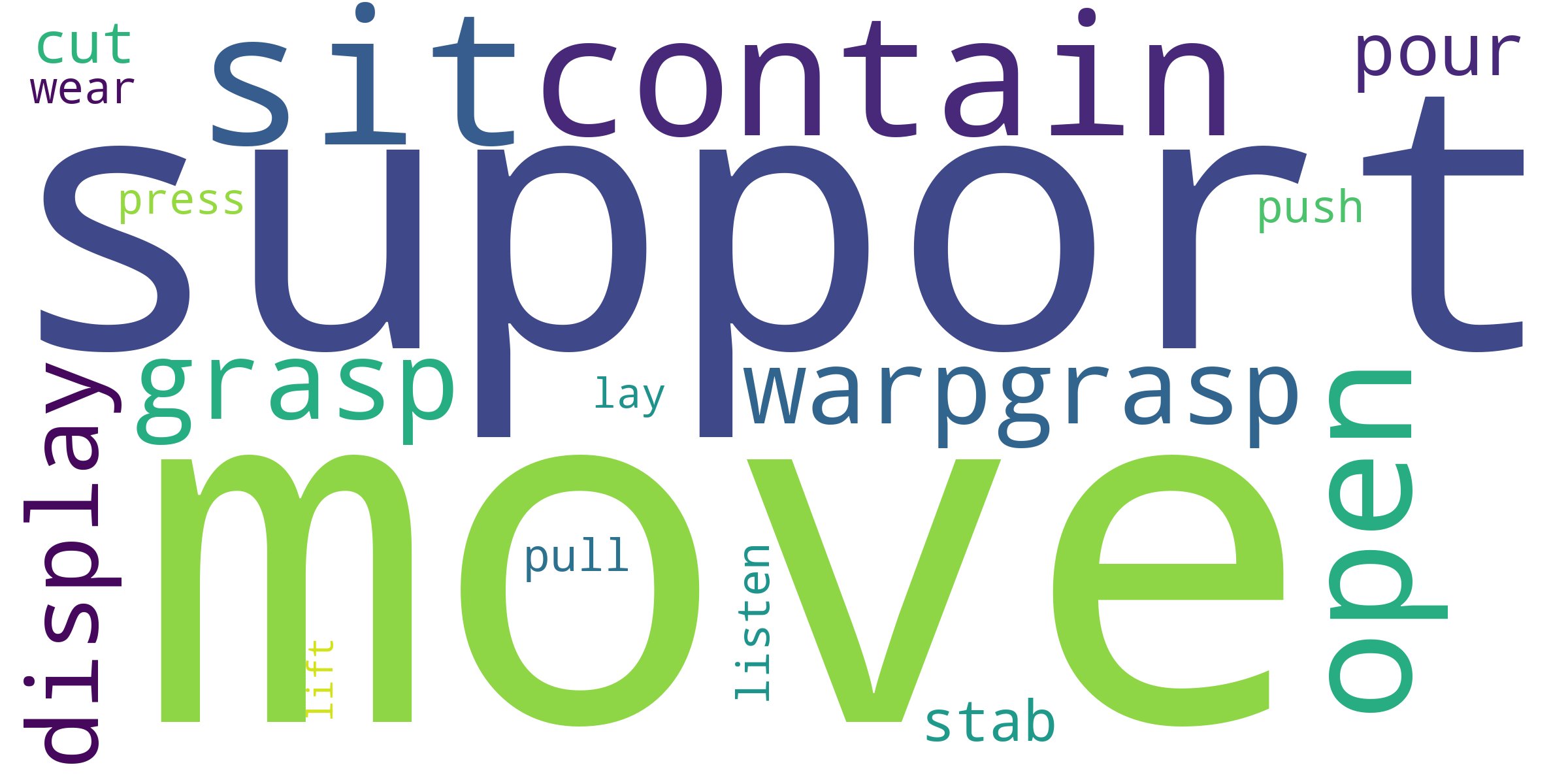}
        \caption{Instructions}
    \end{subfigure}
    \hfill
    \begin{subfigure}{0.325\textwidth}
        \centering
        \includegraphics[width=\textwidth]{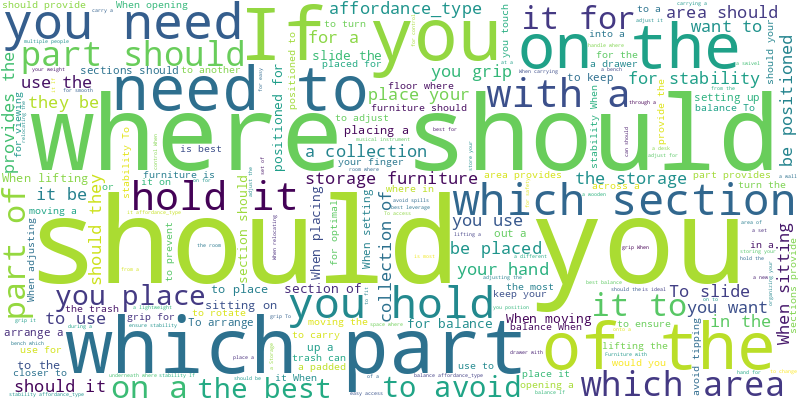}
        \caption{Affordance Types}
    \end{subfigure}
    \hfill
    \begin{subfigure}{0.325\textwidth}
        \centering
        \includegraphics[width=\textwidth]{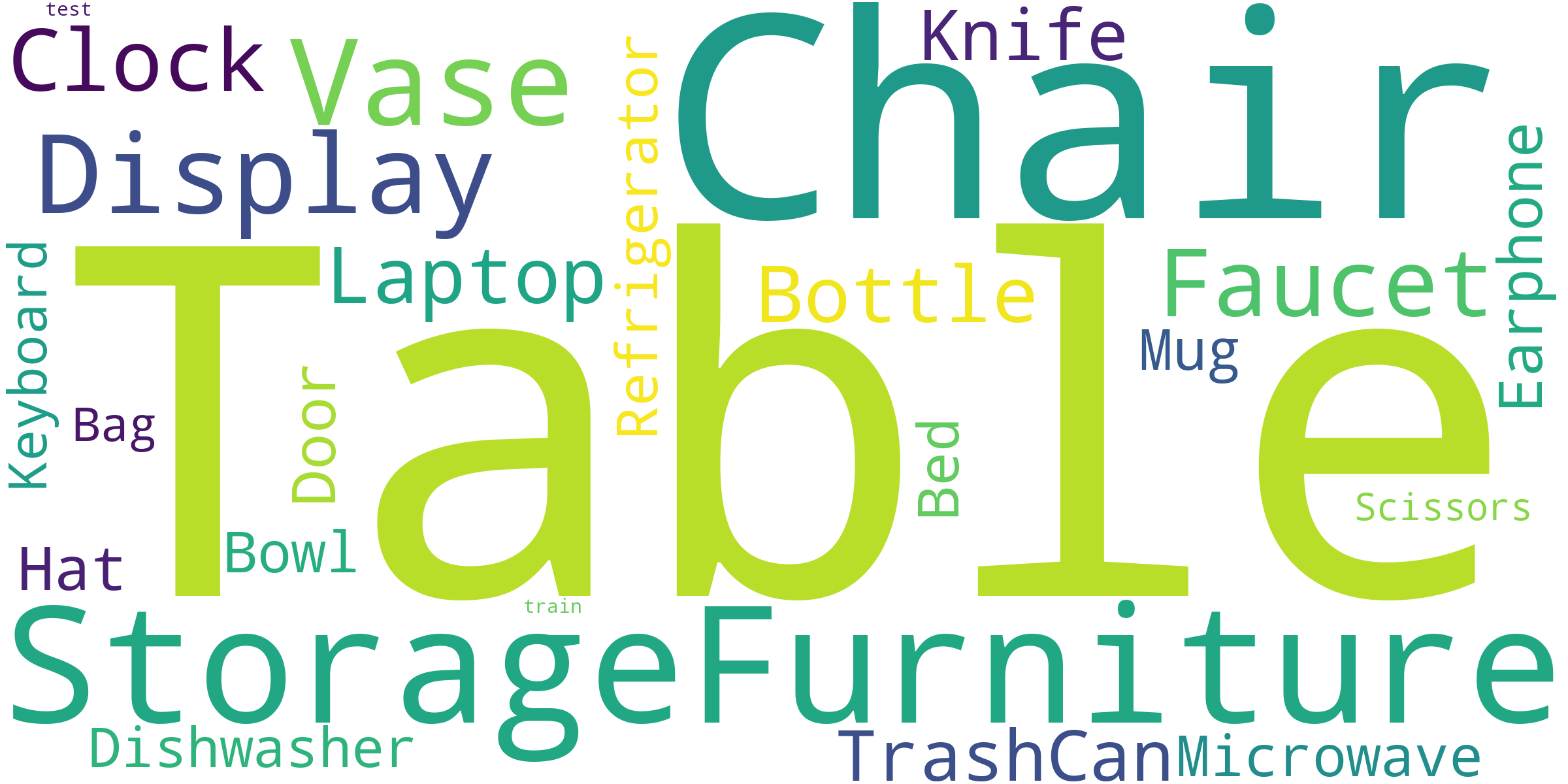}
        \caption{Object Categories}
    \end{subfigure}
    
    \caption{Word clouds of (a) instructions, (b) affordance types, and (c) object categories.}
    \label{ciyun}
\end{figure*}

Our dataset comprises 162,386 instruction-point cloud pairs in the single affordance segmentation setting and 20,847 pairs in the sequential affordance segmentation setting, making up a total of 18,371 point cloud instances across 23 object categories. We have visualized word clouds for the object categories, affordance types, and the introduction in our dataset, demonstrating the richness of our dataset. Fig. \ref{length-que} shows the distribution of the generated question length and Fig. \ref{ciyun} shows the words cloud of the generated instructions.
% \begin{figure*}[!h]
% \centering
% \includegraphics[width=1\linewidth]{img/sup7.png}

% \label{sup7}
% \end{figure*} 
% \begin{figure*}[!h]
% \centering
% \includegraphics[width=1\linewidth]{img/sup8.png}

% \label{sup8}
% \end{figure*} 

% Add actual content here

\begin{figure*}[!h]
\centering
\includegraphics[width=1\linewidth]{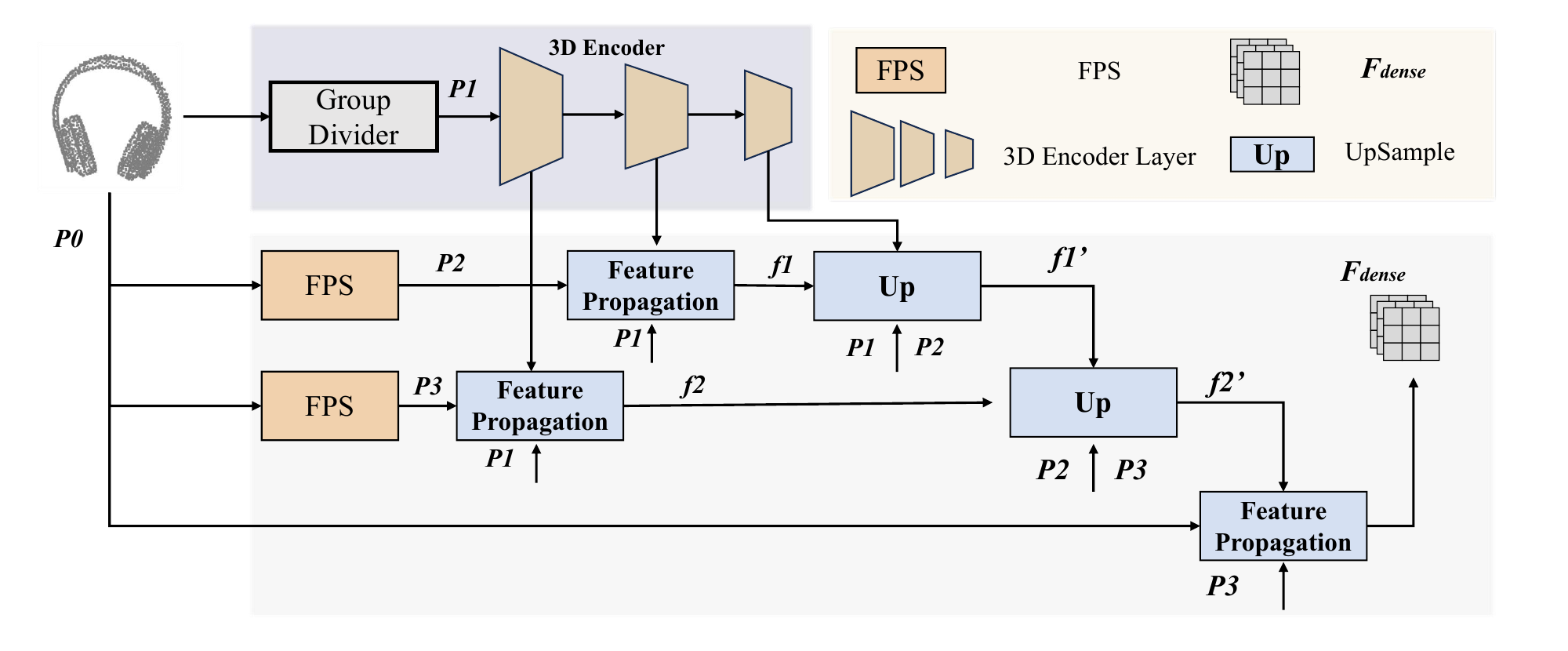}
\caption{\textbf{Upsample.} We show the details of the Upsample method.Specifically, we use FPS and feature propagation to get the hierarchical point cloud features and upsample the sparse features into dense features}

\label{UPSA}
\end{figure*} 

\section{Detailed Methodology}
\label{sec:detailed_methodology}
\subsection{Point Upsample}

 Fig. \ref{UPSA} shows the detailed method of our upsample ways. Here, we adopt Uni3D\citep{zhang2023uni3d} as our 3D encoder, which is capable of encoding point clouds to obtain semantically rich point cloud features $Fp_{sparse}$. To perform dense prediction tasks, we need to obtain finer-grained point cloud features from the sparse point cloud features $Fp_{sparse}$. Following PointNet++\citep{qi2017pointnet++} and DGCNN\citep{wang2019dynamic}, we utilized their feature propagation techniques to build our feature propagation process. Specifically, we extract features from the intermediate layers of the 3D encoder and use Farthest Point Sampling (FPS) to sample different numbers of points from the point cloud:
 \begin{equation}
     P2 = FPS(X_{point}),
 \end{equation}
  \begin{equation}
P3 = FPS(X_{point}).
 \end{equation}

Feature propagation is then applied to obtain features $f1$ and $f2$:

 \begin{equation}
    f1 = FP(P1,P2,H8),
 \end{equation}
  \begin{equation}
    f2 = FP(P1,P3,H4).
 \end{equation}

Next, we follow DGCNN\citep{wang2019dynamic} using its upsample technique to obtain $f1^{'}$ and $f2^{'}$:
 \begin{equation}
f1^{'} = Up(P2,f1,P2.H8),
 \end{equation}
  \begin{equation}
f2^{'} = Up(P3,f2,P2,f1^{'})
 \end{equation}

\subsection{Mutil-Object }
Here, we describe the detailed implementation of the Mutil-Object affordance reasoning. During training, we add $<SEG>$ token after the object name, and our MLLM will learn the paragrim and when it needs to generate text output, it will also follow this paragtim. We extract the nouns from the output of the language model, we then use the order of these nouns as the order of the model affordance reasoning. This is justified because the large language model is able to output the correct text sequence.

\newpage
\section{More Visualization Results}
\begin{figure*}[htbp]
    \centering
    \includegraphics[width=1\linewidth]{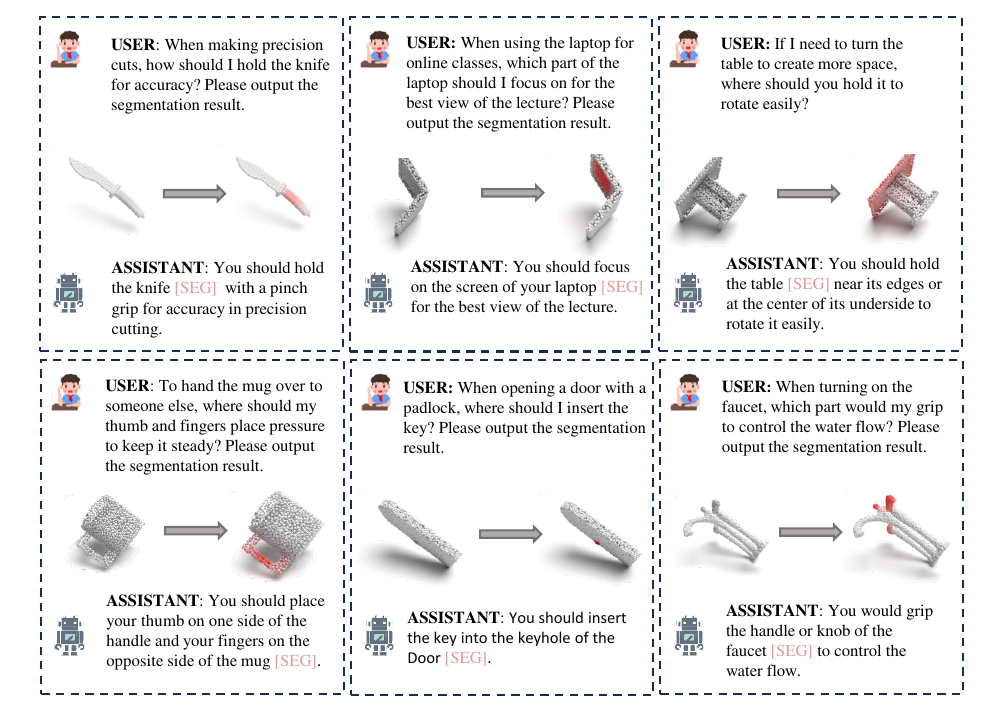}
    \caption{More Qualitative Visualization Results of Sequential Affordance Reasoning}
    \label{table1}
\end{figure*}

\begin{figure*}[htbp]
    \centering
    \includegraphics[width=1\linewidth]{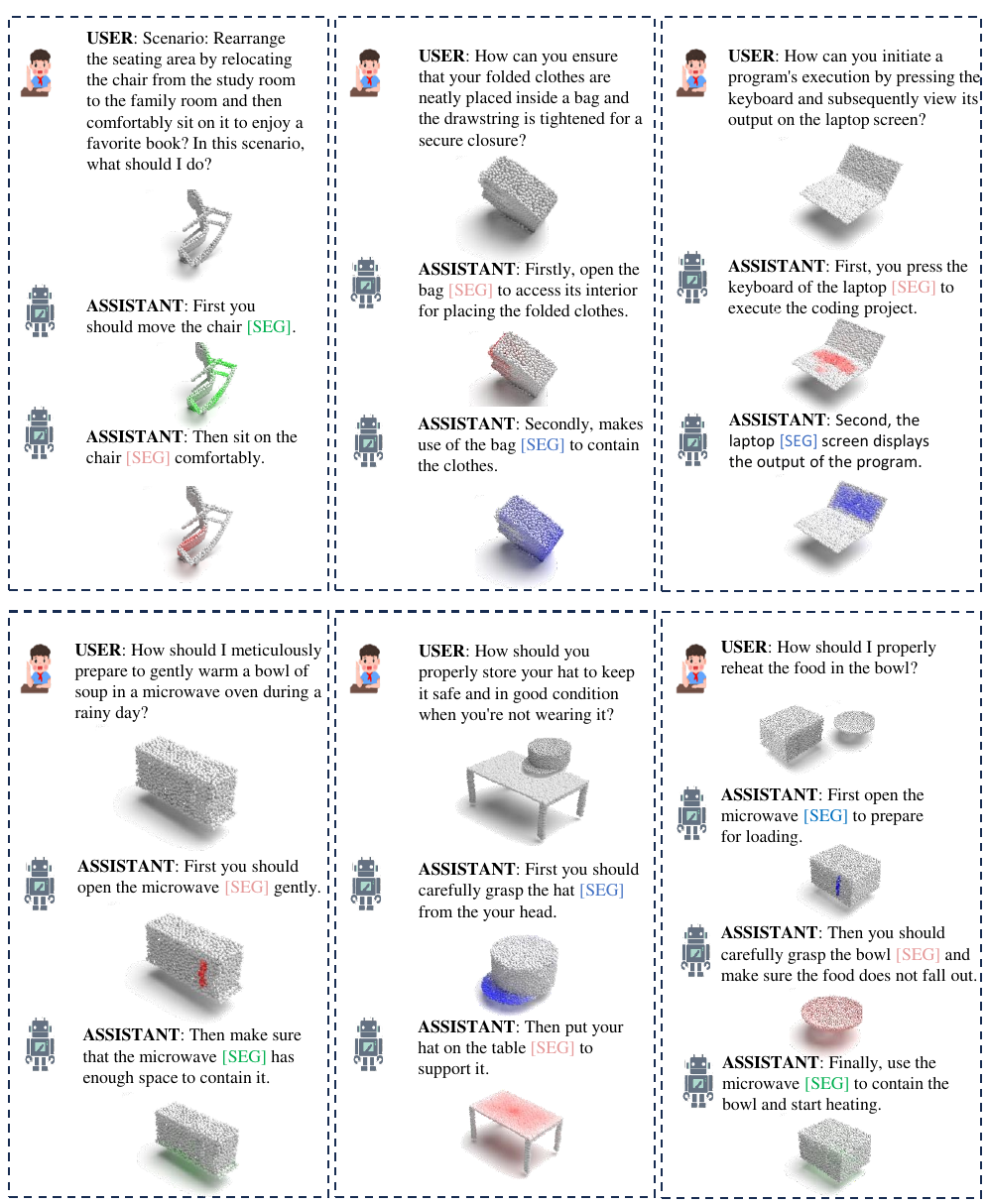}
    \caption{More Qualitative Visualization Results of Sequential Affordance Reasoning}
    \label{table2}
\end{figure*}

Here, we show more qualitative results of our model. Fig. \ref{table1} shows the additional visualization results of Single Affordance Reasoging, and Fig. \ref{table2} shows the additional visualization results of 

\section{Comparison with the Previous SOTA Method}
We show the qualitative comparison with LASO\cite{li2024laso}: obviously, when the question is complex, LASO\cite{li2024laso} failed to understand the intention and do not perform well in connecting the affordance with the intention.
\begin{figure*}[t]
    \centering
    \includegraphics[width=1\linewidth]{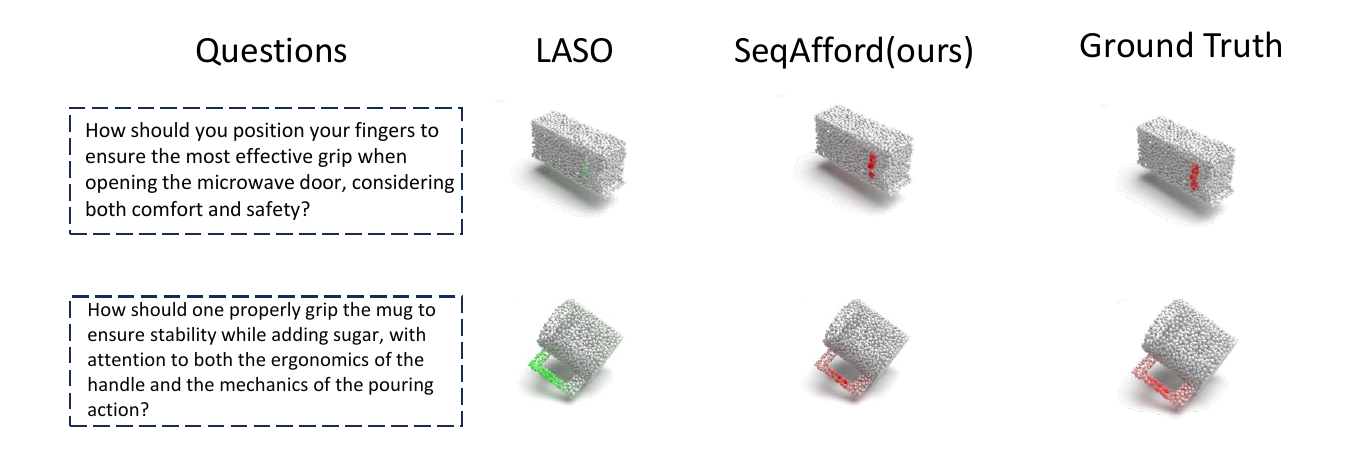}
    \caption{Comparison with the Previous SOTA Method}
    \label{com}
\end{figure*}

\end{document}